\documentclass[10pt,journal,compsoc]{IEEEtran}
\ifCLASSOPTIONcompsoc
  \usepackage[nocompress]{cite}
\else
  \usepackage{cite}
\fi

\ifCLASSINFOpdf
\else
\fi

\usepackage{amsthm}
\usepackage{amsmath,amsfonts}
\usepackage{algorithmic}
\usepackage{algorithm}
\usepackage{array}
\usepackage[caption=false,font=normalsize,labelfont=sf,textfont=sf]{subfig}
\usepackage{textcomp}
\usepackage{stfloats}
\usepackage[hyphens]{url}
\usepackage{breakurl}
\usepackage{verbatim}
\usepackage{graphicx}
\usepackage{cite}
\usepackage{multirow}
\usepackage{vcell}
\usepackage{makecell}
\usepackage[normalem]{ulem}
\usepackage{color}
\usepackage{ragged2e}
\usepackage{tabularray}
\usepackage{arydshln}
\usepackage{hhline}

\usepackage{pifont}

\usepackage{enumitem}
\setlist[itemize]{leftmargin=15pt}

\theoremstyle{definition}
\newtheorem{definition}{\textbf{Definition}}

\begin{document}

\title{Time-aware Metapath Feature Augmentation for Ponzi Detection in Ethereum}

\author{Chengxiang Jin, Jiajun Zhou, Jie Jin, Jiajing Wu,~\IEEEmembership{Senior Member,~IEEE,} Qi Xuan,~\IEEEmembership{Senior Member,~IEEE}
\IEEEcompsocitemizethanks{
\IEEEcompsocthanksitem This work was supported in part by the Key R\&D Program of Zhejiang under Grants 2022C01018 and 2024C01025, by the National Natural Science Foundation of China under Grants 62103374, U21B2001 and 61973273 and by the Key R\&D Projects in Zhejiang under Grant 2021C01117. (Corresponding author: Jiajun Zhou.)
\IEEEcompsocthanksitem C. Jin, J. Jin are with the Institute of Cyberspace Security, College of Information Engineering, Zhejiang University of Technology, Hangzhou 310023, China, with the Binjiang Cyberspace Security Institute of ZJUT, Hangzhou 310056, China. E-mail:\{jincxiang, 2112003197\}@zjut.edu.cn.
\IEEEcompsocthanksitem J. Zhou are with the Institute of Cyberspace Security, College of Computer Science and Technology, Zhejiang University of Technology, Hangzhou 310023, China, with the Binjiang Cyberspace Security Institute of ZJUT, Hangzhou 310056, China. E-mail: jjzhou@zjut.edu.cn.
\IEEEcompsocthanksitem J. Wu is with the School of Computer Science and Engineering, Sun Yat-sen University, Guangzhou 510006, China. E-mail: wujiajing@mail.sysu.edu.cn.
\IEEEcompsocthanksitem Q. Xuan is with the Institute of Cyberspace Security, College of Information Engineering, Zhejiang University of Technology, Hangzhou 310023, China, with the Binjiang Cyberspace Security Institute of ZJUT, Hangzhou 310056, China. E-mail: xuanqi@zjut.edu.cn.
}
}

\markboth{Journal of \LaTeX\ Class Files,~Vol.~14, No.~8, August~2015}%
{Shell \MakeLowercase{\textit{et al.}}: Bare Demo of IEEEtran.cls for Computer Society Journals}

\IEEEtitleabstractindextext{%
\begin{abstract}
  \justifying
  With the development of Web 3.0 which emphasizes decentralization, blockchain technology ushers in its revolution and also brings numerous challenges, particularly in the field of cryptocurrency. Recently, a large number of criminal behaviors continuously emerge on blockchain, such as Ponzi schemes and phishing scams, which severely endanger decentralized finance. 
  Existing graph-based abnormal behavior detection methods on blockchain usually focus on constructing homogeneous transaction graphs without distinguishing the heterogeneity of nodes and edges, resulting in partial loss of transaction pattern information. 
  Although existing heterogeneous modeling methods can depict richer information through metapaths, the extracted metapaths generally neglect temporal dependencies between entities and do not reflect real behavior.
  In this paper, we introduce Time-aware Metapath Feature Augmentation (\emph{TMFAug}) as a plug-and-play module to capture the real metapath-based transaction patterns during Ponzi scheme detection on Ethereum.
  The proposed module can be adaptively combined with existing graph-based Ponzi detection methods.
  Extensive experimental results show that our \emph{TMFAug} can help existing Ponzi detection methods achieve significant performance improvements on the Ethereum dataset, indicating the effectiveness of heterogeneous temporal information for Ponzi scheme detection.
\end{abstract}
}
\maketitle

\IEEEdisplaynontitleabstractindextext
\IEEEpeerreviewmaketitle

\section{Introduction}
\IEEEPARstart{B}{lockchain} technology has been developing rapidly in recent years and gradually gaining public attention.
Blockchain~\cite{li2020survey} is a peer-to-peer network system based on technologies such as cryptography~\cite{stinson2005cryptography} and consensus mechanisms~\cite{lashkari2021comprehensive} to create and store huge transaction information. At present, the biggest application scenario of blockchain is cryptocurrency. For example, the initial ``Bitcoin"~\cite{nakamoto2008bitcoin} also represents the birth of blockchain. 
As cryptocurrencies continue to evolve, smart contracts~\cite{buterin2014next} bring blockchain 2.0, also known as Ethereum~\cite{wood2014ethereum}.
Unlike Bitcoin, which prefers a peer-to-peer electronic cash system, Ethereum is a platform for decentralized applications and allows anyone to create and execute smart contracts. 
The smart contract~\cite{wang2019blockchain}, accompanying Ethereum, is understood as a program on the blockchain that operates when the starting conditions are met.
Since smart contracts operate on publicly accessible code and are immutable, it is possible to carry out secure transactions without third-party endorsement.

The Ponzi scheme~\cite{artzrouni2009mathematics} is a form of fraud that benefits from a poor return on investment to the victim. Traditional Ponzi schemes have existed only offline, but they have gradually taken on an online form as the spotlight of money flows has shifted online. 
Cryptocurrency is trusted by the public for its security. Due to the transparency and immutability of smart contracts on Ethereum, people tend to be less vigilant, which makes it easier for Ponzi schemes to execute. 
According to recent reports\footnote[1]{\url{https://www.sec.gov/news/press-release/2022-134}}, the SEC revealed that the Forsage smart contract platform~\cite{kell2021forsage} is a large fraudulent pyramid scheme that marketed fraudulent products to investors, involving a total of \$300 million and millions of victims.
This undermines the public's trust in cryptocurrencies.
Therefore, there is an urgent need to understand the behavior of Ponzi schemes and detect them from cryptocurrency platforms, further maintaining the stability of the investment markets.

\IEEEpubidadjcol
Existing Ponzi scheme detection methods based on graph analytics generally rely on homogeneous graph modeling~\cite{macpherson2011survey,Ethident,jin2023ethereum} due to their simplicity. However, real transactions on Ethereum generally involve different types of interactions between different types of accounts, which will be neglected during homogeneous modeling. Therefore, it is tough to reflect the complexity and diversity in real transactions simply by homogeneous graphs.
Meanwhile, heterogeneous graph~\cite{shi2016survey} is a widely used technique to model complex interactions, which can preserve the semantic information of interactions to the greatest extent.
Typical heterogeneous techniques generally employ the metapath structure~\cite{sun2011pathsim} to capture the specific interaction patterns.
However, several approaches~\cite{dong2017metapath2vec,shi2018heterogeneous} are obsessed with searching for the best metapaths, which is time-consuming and difficult to define the best.
In addition, the timestamp information will be neglected when extracting metapaths, resulting in an inappropriate afterthought based on timeless metapaths.
Finally, compared with homogeneous methods, heterogeneous methods ensure fine granularity of information, but also bring operational complexity.

The aforementioned issues motivate us to improve Ponzi detection by integrating heterogeneous information and time information into homogeneous methods.
In this paper, we propose Time-aware Metapath Feature Augmentation (\emph{TMFAug}) as a generic module to capture the temporal behavior patterns hidden in Ethereum interaction graph.
Specifically, \emph{TMFAug} first extracts the time-aware metapath features on an auxiliary heterogeneous graph where the coordinated transaction and contract call information are contained, and then proposes the concept of symbiotic relationship to reduce the impact of repetitive metapaths during information aggregation, finally aggregates these heterogeneous features associated with temporal behavior patterns to corresponding account nodes in the homogeneous graph where the Ponzi detection methods are performed.
Our proposed module allows for improving the performance of existing Ponzi detection methods through feature augmentation without adjusting them.
The main contributions of this work are summarized as follows:
\begin{itemize}
  \item We propose time-aware metapaths that impose timestamp constraints on timeless metapaths to ensure capturing temporal account interaction patterns. In addition, we propose symbiotic relationship and behavioral refinement criteria for metapaths that can capture fine-grained interaction patterns while alleviating information redundancy.
  \item We propose a generic time-aware metapath feature augmentation module, named \textit{TMFAug}, which allows for aggregating heterogeneous features associated with temporal behavior patterns to homogeneous transaction graphs, further improving the performance of existing Ponzi detection methods. 
  To the best of our knowledge, the application of heterogeneous methods to blockchain data mining is scarce, and our work earlier explored the heterogeneous strategies for Ethereum Ponzi detection.
  \item Extensive experiments on the Ethereum dataset demonstrate that the \textit{TMFAug} module can effectively improve the performance of multiple existing Ponzi detection methods. 
  Moreover, the generic compatibility of \textit{TMFAug} also suggests that temporal and heterogeneous behavior pattern information can benefit Ponzi scheme detection in Ethereum.
\end{itemize}

The remaining portions of this paper are summarized below. 
Sec.~\ref{sec:related work} reviews the related work in graph representation learning and graph-based Ponzi detection methods. 
Sec.~\ref{sec: graph} describes the details of constructing an account interaction graph on Ethereum.
Sec.~\ref{sec: method} introduces the details of the proposed \textit{TMFAug} module.
The experimental settings and analysis are presented in Sec.~\ref{sec:Experiment}. 
Finally, we conclude this paper and prospect future work in Sec.~\ref{sec:conclusion}.

\section{Related Work} \label{sec:related work}
In this section, we review several established graph-related methods and introduce their applications in Ethereum Ponzi scheme detection. 

\subsection{Graph Representation Learning} \label{sec:RW-B}
Graph representation learning aims to generate low-dimensional vectors that can capture graph properties and structure and use them for downstream tasks such as node classification~\cite{gong2023neighborhood,gong2023clarify}, link prediction~\cite{fu2018link,yu2019target}, community detection~\cite{zhou2021robustecd,su2022comprehensive} and graph classification~\cite{xuan2019subgraph,zhou2020m}.
Early related works focus on learning topological embeddings, classical ones include Deepwalk, Line, and Node2Vec.
Deepwalk~\cite{perozzi2014deepwalk} first generates node sequences by random walks, then regards them as sentences and feeds into language model for learning node representations.
On this basis, Node2Vec~\cite{grover2016node2vec} introduces two parameters to control the trade-off between the breadth-first search and depth-first search of random walks in order to better capture the topological characteristics.
Line~\cite{tang2015line} is applicable to various kinds of networks as well as large networks and takes into account the first-order and second-order similarities between nodes.
The above methods are task-agnostic and only focus on the topology of the graph, which generally leads to mediocre performance on downstream tasks.
Graph neural networks (GNNs) simultaneously learn the attribute features and topological features of graphs, which brings graph representation learning to a new level.
As a pioneer in GNNs, GCN~\cite{kipf2016semi} is a semi-supervised learning method that first applies the convolutional algorithm in the graph. 
Following GCN, GAT~\cite{velickovic2017graph} considers the importance of neighbors and introduces self-attention mechanism to assign an aggregated weight to each neighbor.
On the other hand, SAGE~\cite{hamilton2017inductive} uses the sampling strategy to transform the full graph training into subgraph mini-batch training, which greatly improves the scalability of GNN on large graph computation.
Additionally, GIN\cite{xu2018powerful} defines the graph neural network expressivity problem and designs an injective neighbor aggregation function in a simple setup formalism.

\begin{table}
  \centering
  \renewcommand\arraystretch{1.2}
  \caption{Notations and Descriptions.}
  \label{tab: notation}
  \resizebox{\linewidth}{!}{
  \begin{tabular}{cr} 
  \hline\hline
  Notation                      & Description                           \\ 
  \hline
  $G_{hom}$                     & Homogeneous graph          \\
  $G_{het}$                     & Heterogeneous graph    \\
  $\boldsymbol{x}$              & Account feature                       \\
  $\hat{\boldsymbol{x}}$        & Account feature with \emph{TMFAug}       \\
  $P $                          & Time-aware Metapath    \\
  $\mathcal{P}$                 & Super metapath                             \\
  $\mathcal{P}_i$               & The $i$-th category of super metapath \\
  $\overrightarrow{\mathcal{V}}(\mathcal{P})$   & The node sequence of the super metapath   \\
  $\overrightarrow{\mathcal{R}_t}(\mathcal{P})$   & The relation types sequence of the super metapath   \\
  $\mathcal{M}$                 & The set of super metapath with same head node  \\
  $\mathcal{M}_\ast$            & The set of super metapath with same target node  \\
  $\omega$                      & The important factor of the super metapath       \\
  $\hat{\omega}$                & The normalized factor of the super metapath                 \\
  \hline\hline
  \end{tabular}}
\end{table}

\subsection{Ponzi Schemes Detection in Ethereum} \label{sec:RW-A}
In 1919, a speculative businessman named Charles Ponzi made fictitious investments in a company with the intention of luring other individuals into the scam by promising new investors a quick return on their initial investment.
Consequently, this type of fraud is known as a Ponzi scheme.
Traditional Ponzi schemes were offline, while recently there has been a shift to online.
Moore et al.~\cite{moore2012postmodern} described a type of high-yield investment program (HYIPs), an online Ponzi scheme. 
They argued that some investors know well the fraudulent nature of these sites, but they believe they can gain partial benefit by investing early in the Ponzi and withdrawing their funds before the Ponzi scheme collapses. 
This idea makes it easy for Ponzi fraudsters to obtain the initial investment.
As time goes by, HYIPs are also active in cryptocurrencies. 
Vasek et al.~\cite{vasek2015there} analyzed the behavior of Ponzi schemes involving Bitcoin in HYIPs, which they called Bitcoin-only HYIPs.
They introduced a variety of fraudulent methods, such as one part claiming to be a legitimate investment vehicle, another part claiming to be an online bitcoin wallet that offers high daily returns, etc.

Analysis of HYIPs has revealed that a number of traditional Ponzi schemes have gradually emerged in cryptocurrencies, generating unique means of deception. 
For example, due to the non-interruptible nature of smart contracts, a lack of attention to contract content can lead to the failure to withdraw funds before a Ponzi scheme collapses, thus amplifying the losses of less sophisticated investors.
Nowadays, multiple detection methods have been proposed to mitigate the damage caused by Ponzi schemes.

\subsubsection{Detection based on Contract Codes}
The code of a smart contract defines the functionality of the contract account, enabling analysis to determine whether it is a Ponzi account.
Chen et al.~\cite{chen2018detecting} extracted features from user accounts and smart contract opcodes, and fed them into a downstream machine learning classifier to detect Ponzi schemes.
To enhance the synergy between feature engineering and machine learning, Fan et al.~\cite{fan2020expose} employed the concept of ordered augmentation to train a model for detecting Ponzi schemes. 
Bartoletti et al.~\cite{bartoletti2020dissecting} explored the significance of temporal behavior and designed time-dependent features for Ponzi detection.
Due to the susceptibility of machine learning models to evasion techniques, Chen et al.~\cite{chen2021sadponzi} proposed a semantics-aware detection method called SADPonzi. This method analyzes the contractual rules of a Ponzi scheme and heuristically generates semantic paths based on the teEther~\cite{krupp2018teether} vulnerability detection algorithm. 
Given the inherent challenges in capturing the structural and semantic features of source code behavior through feature engineering, Chen et al.~\cite{chen2021improving} employed graph embedding techniques to automatically acquire highly expressive code features.

\subsubsection{Detection based on Transactions}
Contract code features provide insights into the inherent characteristics of smart contracts, however, 
fraudsters possess the ability to intentionally evade fixed patterns, thereby rendering detection more challenging.
Fortunately, 
transaction features describe the actual interaction patterns generated by Ponzi accounts, from which the illegality of transactions can be effectively identified.
With the development of machine learning, Zhang et al.~\cite{zhang2021detecting} proposed a new method for Ethereum Ponzi scheme detection based on an improved LightGBM algorithm, which can alleviate the imbalance problem of Ponzi data.
Graph neural networks are capable of learning both account features and graph structural features, Yu et al.~\cite{yu2021ponzi} constructed transaction graphs and utilized GCN~\cite{kipf2016semi} to characterize the account features for Ponzi detection. 
Liu et al.~\cite{liu2022fa} first performed feature enhancement to aggregate high order information and then utilized GCN to learn account transaction features.
Timestamp information is very important for modeling temporal transaction behavior.
Wu et al.~\cite{wu2020phishers} proposed Trans2Vec, a random walk-based approach specifically designed for transactional networks, which takes into account both transaction amounts and timestamps.
Jin et al.~\cite{jin2022dual} used both transaction information and code information to propose a dual-view Ponzi scheme early warning framework, while introducing a temporal evolution augmentation strategy to augment the limited Ponzi dataset.

However, the above-mentioned Ponzi detection methods suffer from several shortcomings.
The above methods rely on homogeneous graphs, and neglect timestamps information and the types of nodes and edges, which makes it difficult to capture more complex and temporal behavior patterns.
Jin et al.~\cite{HFAug} explored this problem on Ethereum Ponzi detection, and proposed to fuse heterogeneous information in metapaths and their subsets, further enhancing existing Ponzi detection methods.
However, they neglected the timestamp information so that the constructed metapaths can not reflect the real behavior patterns.
Therefore, in this paper, we focus more on capturing certain temporal and structural information from metapaths, and work on the integration of heterogeneous and temporal information into homogeneous approaches for Ponzi scheme detection.
  
\begin{figure*}
  \centering
\includegraphics[width=0.9\linewidth]{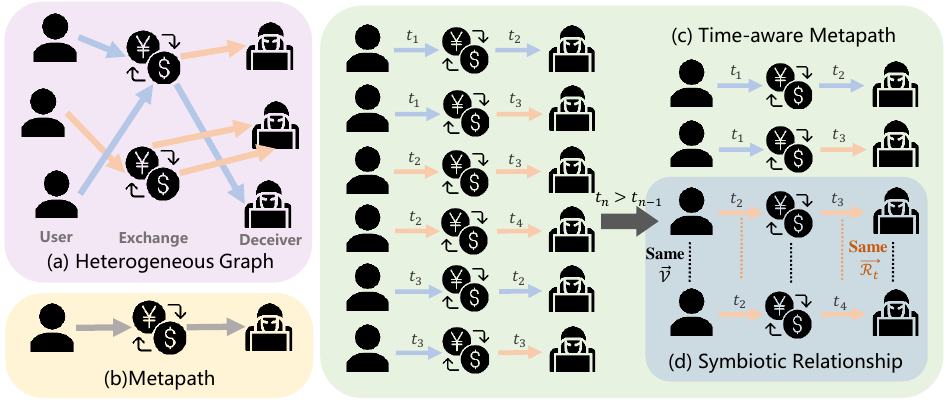}
\caption{Schematic illustration of heterogeneous graphs, metapath, time-aware metapath, and symbiotic relationship.}
\label{fig:preliminaries}
\end{figure*} 

\section{Preliminaries and Terminologies} \label{sec: pre}

In this section, we will introduce some concepts related to heterogeneous graph, including the definitions of heterogeneous graph itself, metapath, our proposed time-aware metapath, and the symbiotic relationship of metapaths.
Fig.~\ref{fig:preliminaries} provides a schematic illustration to further complement the description of the aforementioned concepts.

\begin{definition}[\textbf{Heterogeneous Graph}]
  A heterogeneous graph is defined as $G_\textit{het} = (V,E,\mathcal{T}_V,\mathcal{T}_E)$, where $V$ and $E$ represent the sets of nodes and edges respectively, $\mathcal{T}_V$ and $\mathcal{T}_E$ represent the sets of node types and edge types respectively, whereby each node (or edge) possesses a type mapping function denoted as $V\rightarrow  \mathcal{T}_V$ (or $E\rightarrow  \mathcal{T}_E$) respectively.
  A heterogeneous graph encompasses multiple types of nodes or edges, thereby satisfying $|\mathcal{T}_V| + |\mathcal{T}_E| > 2$.
\end{definition}
\begin{definition}[\textbf{Metapath}]
  In heterogeneous graphs, metapath contains a sequence of relations defined between different types of nodes, such as $v_{1} \stackrel{r_{1}}{\longrightarrow} v_{2} \stackrel{r_{2}}{\longrightarrow} \cdots \stackrel{r_{l}}{\longrightarrow} v_{l+1}$, which specifies a composite connection between node $v_1$ and $v_{l+1}$ with $r=r_{1} \circ r_{2} \circ \cdots \circ r_{l}$, where $\circ$ denotes the composition operation on relations.
\end{definition}

In this paper, we propose the concept of time-aware metapath, which incorporates temporal dependencies in the sequence of relations to capture more realistic and temporally influenced node interaction patterns.
\begin{definition}[\textbf{Time-aware Metapath}]
  For heterogeneous graphs containing temporal relations, time-aware metapath can be defined as $v_{1} \xrightarrow[t_1]{r_1} v_{2} \xrightarrow[t_2]{r_2} \cdots \xrightarrow[t_l]{r_l} v_{l+1}$, where $t$ represents the timestamp information of the corresponding edge and satisfies $t_{1} < t_2<\cdots <t_l$.
\end{definition}

We also introduce a new concept of symbiotic relationship to represent the full consistency of two time-aware metapaths after ignoring temporal information.
\begin{definition}[\textbf{Symbiotic Relationship}]
  For time-aware metapaths $(P_1, P_2, \cdots, P_i)$, they are symbiotic if and only if they have the same node sequence and relation type sequence, i.e.,
  \begin{equation}
    \begin{array}{c}
      P_1 \cong P_2 \cong \cdots \cong P_i \vspace{1ex} \quad 
      \text{s.t.} \\ 
      \overrightarrow{\mathcal{V}} (P_1) =\overrightarrow{\mathcal{V}} (P_2) = \cdots = \overrightarrow{\mathcal{V}} (P_i) \\ 
      \overrightarrow{\mathcal{R}_t} (P_1) = \overrightarrow{\mathcal{R}_t} (P_2) = \cdots = \overrightarrow{\mathcal{R}_t} (P_i)
    \end{array}
  \end{equation}
  where $\cong$ represents a symbiotic relationship, $\overrightarrow{\mathcal{V}}$ and $\overrightarrow{\mathcal{R}_t}$ represent the node sequence and relation type sequence, respectively.
\end{definition}

\section{Account Interaction Graph Modeling} \label{sec: graph}
In this section, we focus on modeling Ethereum transaction data as account interaction graphs, as well as providing a brief introduction to Ethereum data.

\begin{figure*}
  \centering
\includegraphics[width=\textwidth]{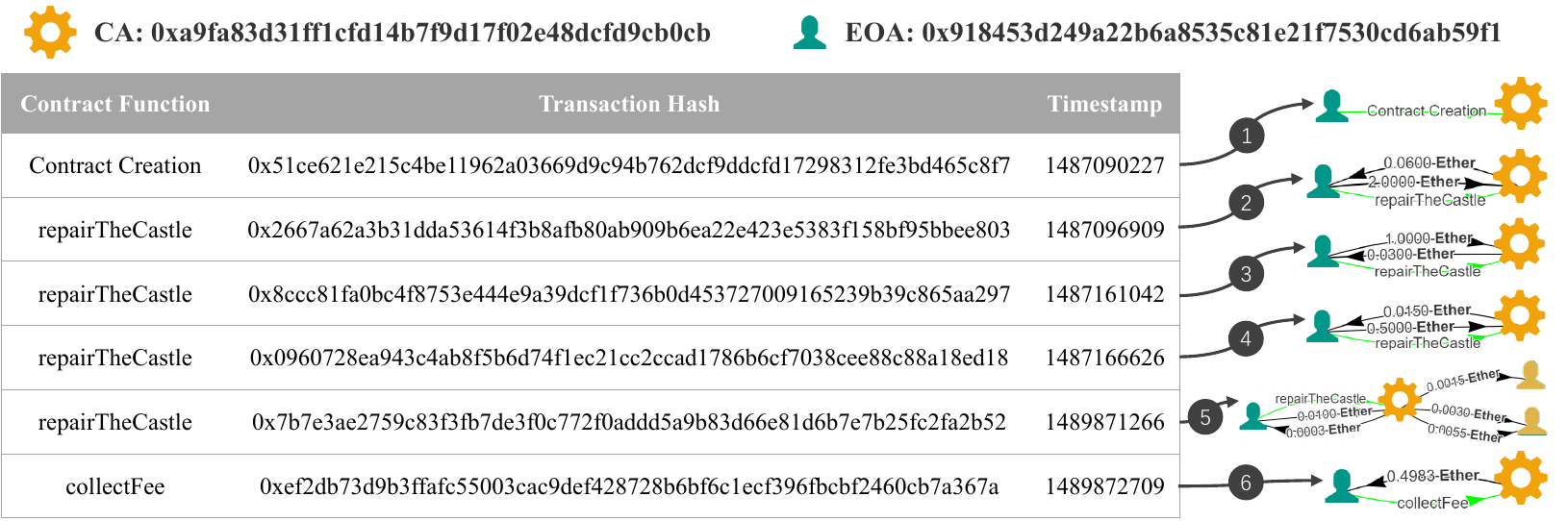}
\caption{An example of a Ponzi scheme disguised as a game.}
\label{fig:transhash}
\end{figure*} 

\subsection{Ethereum Data}
An account in Ethereum is an entity that owns Ether and can be divided into two categories: Externally Owned Account (EOA) and Contract Account (CA)~\cite{wood2014ethereum}.
EOA is controlled by a user with a private key and can initiate transactions on Ethereum, and CA is controlled by smart contract code and can only get triggers to implement the contract function.
There are generally two categories of interactions between Ethereum accounts: transaction and contract call.
The transaction is the process by which an action is initiated by an EOA and received by another EOA, or ultimately back to itself.
Contract calls, which are divided into external calls and internal calls, refer to the process of triggering a smart contract code that can execute many different actions, such as transferring tokens or creating a new contract.

Fig.~\ref{fig:transhash} illustrates a Ponzi contract~\footnote{\url{https://cn.etherscan.com/address/0xa9fa83d31ff1cfd14b7f9d17f-02e48dcfd9cb0cb#code}} in the guise of a game, with a large number of interactions between the CA and an EOA involving calls to different contract functions for different purposes:
\begin{itemize}
  \item \textbf{Contract Creation:} Create a new smart contract with built-in opcodes or keywords.
  \item \textbf{repairTheCastle:} Function customized by the contract owner, open to be called by the investor. When the invested amount exceeds 100 ETH, the excess is returned to the last three investors in the ratio of 55\%, 30\% and 15\%, while the initial investor receives a return of 3\%. Incur a cumulative fee of 3\%.
  \item \textbf{collectFee:} Function customized by the contract owner. The contract owner collects all the accumulated fees.
\end{itemize}
Notably, the process of contract call is accompanied by the flow of Ether. A transaction initiated by an EOA may be accompanied by multiple sub-transactions related to the smart contract at the same time, as shown in transaction \ding{186}. During graph modeling, for complex interaction scenarios, we consider multiple sub-transactions with the same transaction hash as different interaction edges but with the same timestamp. The time-aware metapaths can identify whether they belong to the sub-transactions in the same transaction behavior by comparing the timestamp information.

\begin{figure}
  \centering
  \includegraphics[width=0.8\linewidth]{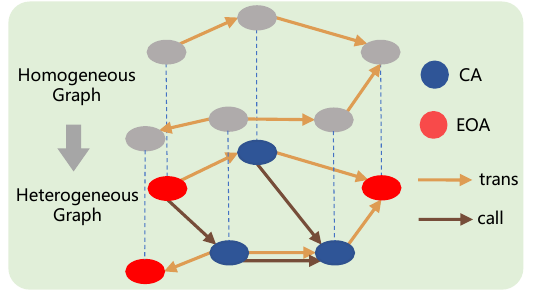}
  \caption{Homogeneous and heterogeneous interaction graph.}
  \label{fig:hom-het}
\end{figure}

\subsection{Graph Modeling on Ethereum}\label{sec: graph-modeling}
The existing Ponzi scheme detection methods mainly use homogeneous transaction graphs, which regard all nodes and edges as the same type, neglecting the complex and temporal interaction information. 
On the contrary, heterogeneous graphs group nodes and edges into various categories, which can represent more complex interaction scenarios and make full use of graph information. 
Considering that most of the existing Ponzi detection methods are only suitable for homogeneous graphs, we consider extracting information from heterogeneous graphs to assist in Ponzi detection on homogeneous graphs.

More formally, we use $G_\textit{hom}=(V, E, Y)$ and $G_\textit{het}=(V_\textit{eoa}, V_\textit{ca}, E_\textit{trans}^t, E_\textit{call}^t, Y)$ to represent the two types of graphs respectively, 
where $V$ represents the set of accounts in the Ethereum data, 
$V_\textit{eoa}$ and $V_\textit{ca}$ represent the sets of EOA and CA respectively,
$E$ represents the set of directed edges constructed from transaction information,
$E_\textit{trans}^t$ and $E_\textit{call}^t$ represent the sets of directed edges with timestamps constructed from transaction information and contract call information respectively,
$Y=\{(v_\textit{i}^\textit{p}, y_\textit{i})\}$ is the label information of known Ponzi accounts. 
Notably, all the known Ponzi schemes we have collected on Ethereum are based on contract accounts.
The nodes of $G_\textit{hom}$ and $G_\textit{het}$ are aligned, as illustrated in Fig.~\ref{fig:hom-het}.
Note that in practice, there may be multiple edges with different timestamps between any two account nodes 
Compared with $G_\textit{hom}$, $G_\textit{het}$ has additional account category information (i.e., EOA and CA), another interactive edge information (i.e., contract call), and timestamps in edges.

During graph modeling, for the homogeneous transaction graph, we neglect timestamps and merge multiple edges between node pairs into a single edge, while attaching two new features to it, representing the number of merged transactions and the total amount of transactions. The final generated homogeneous transaction graph contains 57,130 nodes and 86,602 edges.
For the heterogeneous interaction graph, we take timestamps into account and distinguish between the different node and edge types, classifying them into two categories: 4,616 CA nodes and 52,514 EOA nodes, 984,498 transaction edges and an additional 1,780,781 call edges.
In addition, we crawled 191 labeled Ponzi data from \textit{Xblock}\footnote[1]{\url{http://xblock.pro/ethereum/}}, \textit{Etherscan}\footnote[2]{\url{https://cn.etherscan.com/accounts/label/ponzi}} and other Blockchain platforms.
For all detection methods, we take all the labeled Ponzi accounts as positive samples, as well as the same number of randomly sampled CA as negative samples.
The statistics of the two graphs are shown in Table~\ref{tab: graph}.

\begin{table}
  \centering
  \caption{Statistics of the datasets. $|V|$ and $|E|$ are the total number of nodes and edges respectively. $|V_\textit{ca}|$ and $|V_\textit{eoa}|$ are the number of CA and EOA respectively, $|E_\textit{call}|$ and  $|E_\textit{trans}|$ are the number of call and trans edges respectively, and $|Y|$ is the number of labeled Ponzi accounts.}
  \resizebox{\linewidth}{!}{
  \renewcommand\arraystretch{1.2}
  \begin{tabular}{lccccccc} 
  \hline\hline
      Dataset                        & $|V|$  & $|E|$     & $|V_\textit{ca}|$ & $|V_\textit{eoa}|$ & $|E_\textit{call}|$ & $|E_\textit{trans}|$ & $|Y|$       \\ 
      \hline
      $G_\textit{hom}$   & 57,130 & 86,602    & \multicolumn{4}{c}{$\cdots \quad \text{No label information} \quad \cdots$}                      & 191                   \\
      \hline
      $G_\textit{het}$ & 57,130 & 2,765,279   & 4,616             & 52,514             & 1,780,781               & 984,498              & 191              \\
  \hline\hline
  \end{tabular}}
  \label{tab: graph}
\end{table}  

\subsection{Node Feature Construction}\label{sec: manual-feature}
In this paper, we divide the existing Ponzi detection methods into three categories, namely manual feature engineering, graph embedding and graph neural network.
For Ponzi detection methods based on manual feature engineering and graph neural network, we construct initial features for account nodes in both $G_\textit{hom}$ and $G_\textit{het}$ using 15 manual features proposed in existing methods. 
\begin{figure}
  \centering
  \includegraphics[width=\linewidth]{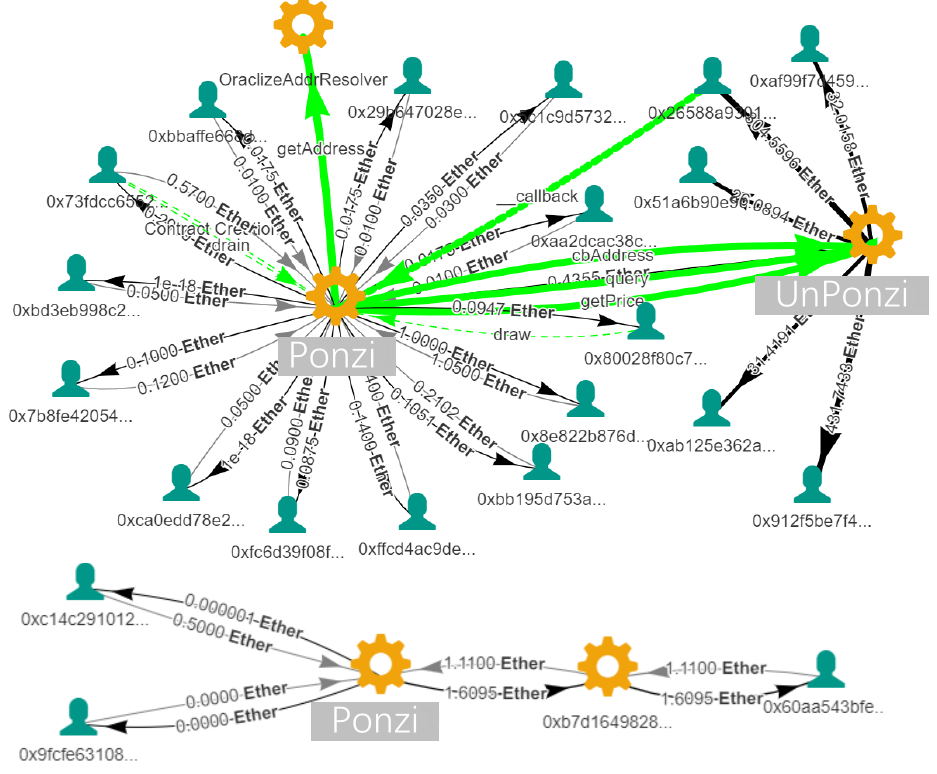}
  \caption{Interaction graphs of real Ponzi schemes in Ethereum.}
  \label{fig:ponzi_inter}
\end{figure}
\begin{itemize}
  \item The income and expenditure of the target account (including total, average, maximum and variance);
  \item The expenditure-income ratio of the target account;
  \item The balance of the target account;
  \item The number of transactions sent and received by the target account;
  \item The investment Gini and return Gini of target account;
  \item The life cycle of the target account.
\end{itemize}
As for methods based on graph embedding, we generate structural embeddings as account node features rather than the predefined manual features. 
After that, the initial feature of arbitrary account node $v_i$ is denoted as follows:
\begin{equation}
  \boldsymbol{x}_{i}=\left\{\begin{array}{ll}
    \left[x_i^1, x_i^2, \cdots,x_i^{15}\right] &\text{\textbf{for} manual feature} \vspace{1ex}\\ 
    \left[x_i^1, x_i^2, \cdots,x_i^{15}\right] &\text{\textbf{for} graph neural network} \vspace{1ex}\\
    \operatorname{GEmb}(G_\textit{hom}, v_\textit{i}) & \text{\textbf{for} graph embedding} 
  \end{array}\right.
\end{equation}
where $x_i$ represents the $i$-th manual feature, and $\operatorname{GEmb}$ represents an arbitrary graph embedding algorithm that generates node embeddings as initial node features.

\section{Time-aware Metapath Feature Augmentation}\label{sec: method}
\begin{figure*}
  \centering
  \includegraphics[width=\textwidth]{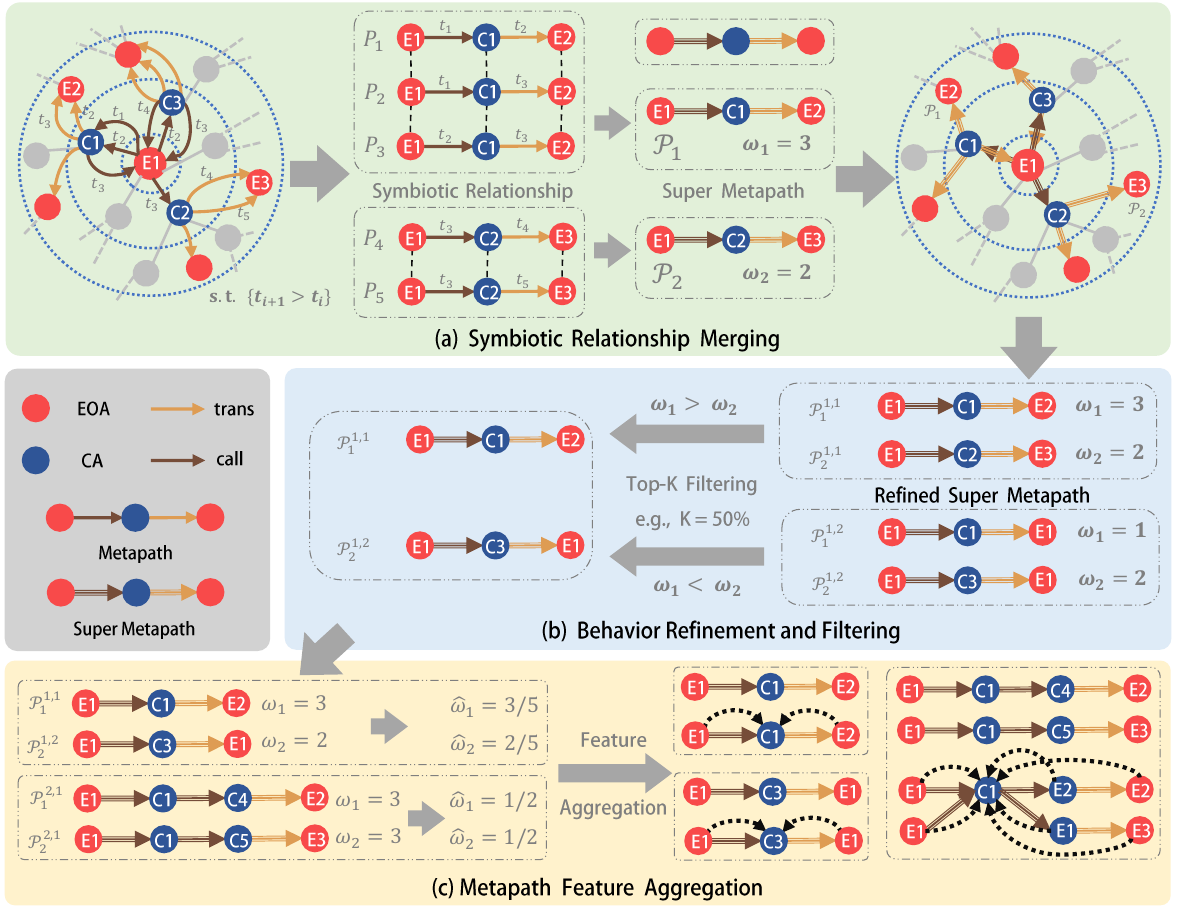}
  \caption{The overall framework of the Time-aware Metapath Feature Augmentation. The complete workflow proceeds as follows: (a) merging symbiotic relations to obtain the super metapaths; (b) refining the super metapaths based on the behavior, and filtering the super metapaths by Top-$K$; (c) aggregating the information along metapaths to the target nodes.}
  \label{fig: method-1}
\end{figure*}

In this section, we introduce the details of \textit{TMFAug}, as schematically depicted in Fig.~\ref{fig: method-1}, which is composed of three components: symbiotic relationship merging, behavior refinement and filtering, and metadata feature aggregation.

\subsection{Time-aware Metapath}
Fig.~\ref{fig:ponzi_inter} shows several representative Ponzi scheme interaction networks in Ethereum. 
The gear icon and portrait icon represent CA and EOA respectively, the black line represents transactions with the specific transaction amount, and the green line represents the contract call with a specific function. 
The CA on the left is a Ponzi account that defrauds investors by engaging in two-way transactions, in which it earns more and pays less.
In addition, Fig.~\ref{fig:transhash} also illustrates that transactions can be triggered by contract calls.
Based on the above observations, we can predefine the following metapaths to characterize the critical behavior patterns present in Ethereum Ponzi schemes:
\begin{equation}\label{eq: notime}
  \begin{array}{l}
    P^{1}_{\times}: \quad EOA_1 \xrightarrow{call} CA_\ast  \xrightarrow{trans} EOA_2 \vspace{1ex}\\
    P^{2}_{\times}: \quad EOA_1 \xrightarrow{call} CA_\ast  \xrightarrow{call} CA_1 \xrightarrow{trans} EOA_2
\end{array}
\end{equation}
The rationality of the above predefined metapaths can be confirmed by statistical evidence. Specifically, we extract all metapaths that meet the above definition from the Ethereum interaction graph constructed in Sec.~\ref{sec: graph-modeling}, using the labeled 382 CA (Ponzi and non-Ponzi) as $CA_\ast$, and ultimately obtain a total of 6,936,126,959 metapaths, of which the $P^{1}_{\times}$ occupy 99.998\%. Furthermore, for all $P^{1}_{\times}$ and $P^{2}_{\times}$ obtained, the number of metapaths in which $CA_\ast$ is a Ponzi account occupies 94.309\%.
The above evidence suggests that Ponzi accounts usually interact with EOA via mode $P^{1}_{\times}$. And $EOA_2$ could be an external investor or the Ponzi contract creator, the former indicates that \emph{trans} is a payback, while the latter indicates that \emph{trans} is a transfer of funds to the defrauder.
In addition, a tiny percentage of Ponzi accounts transfer money via mode $P^{2}_{\times}$, i.e., internal calls. This mode circumvents the regular transaction pattern to some extent, but it is rarely used because too many contract calls consume more gas fees and increase the cost of fraud.

Ponzi schemes reward old investors through new investment income, i.e., two-way transactions do not occur simultaneously, so here we further introduce the time information of transactions to depict the temporal behavior patterns.
Formally, we use the following \textbf{time-aware metapaths} to characterize the temporal behavior patterns:
\begin{equation}\label{eq: p1p2}
  \begin{array}{l}
    P^{1}: EOA \xrightarrow[t_1]{call} CA_\ast  \xrightarrow[t_2]{trans} EOA \vspace{1ex}\\
    P^{2}: EOA \xrightarrow[t_1]{call} CA_\ast  \xrightarrow[t_2]{call} CA \xrightarrow[t_3]{trans} EOA \vspace{1ex}\\
    \quad\quad\quad\quad\quad\quad \text{s.t.} \quad t_1<t_2<t_3 
\end{array}
\end{equation}
Note that the timing constraint $t_1<t_2<t_3$ ensures that the behavior patterns represented by the metapaths are more consistent with the real interaction rules.
Under the timing constraint, the number of effective metapaths is reduced to 3,280,360,935. More specifically, 93.837\% of the time-aware metapaths have a $CA_\ast$ that corresponds to a Ponzi account (93.839\% for $P^1$ and 11.733\% for $P^2$), which also suggests that compared with normal accounts, Ponzi accounts are more inclined to use the interaction patterns defined by the above metapaths to commit fraud.
Additionally, we refer to the metapaths defined in Eq.~(\ref{eq: notime}) as timeless metapaths, which ignores the timing of interactions and is based on an after-the-fact perspective.

\subsection{Symbiotic Relationship Merging}
There are usually frequent interactions between multiple accounts, yielding multiple metapaths with the same sequences of nodes and relationships, but different sequences of timestamps.
Here we consider these metapaths to be symbiotic with each other, and the concept of symbiotic relationship is defined in Sec.~\ref{sec: pre}.
As illustrated in Fig.~\ref{fig: method-1}(a),   $P_1$, $P_2$ and $P_3$ are symbiotic to each other because they have the same sequences of nodes and relationships $E1 \xrightarrow{call} C1 \xrightarrow{trans} E2$, and so do $P_4$ and $P_5$.
Since symbiotic metapaths are numerous and have the same form, we use a super metapath to substitute them.
Specifically, given a set of symbiotic metapaths $\{P_1,P_2,\cdots, P_n \mid P_1 \cong P_2 \cong \cdots \cong P_n \}$, we merge these symbiotic metapaths into a super metapath $\mathcal{P}$.
Intuitively, the number of symbiotic metapaths reflects the account's preference in participating in this behavior to a certain extent.
So we measure the importance of the super metapath via the number of symbiotic metapaths and further assign it an importance factor $\omega = n$.
As illustrated in Fig.~\ref{fig: method-1}(a), $\{P_1,P_2,P_3\}$ (or $\{P_4,P_5\}$) is merged into a super metapath $\mathcal{P}_1$ (or $\mathcal{P}_2$) and assigned an importance factor $\omega_1=3$ (or $\omega_2=2$).

After symbiotic relationship merging, we recount the number of time-aware metapaths (and the proportion of $CA_\ast$ that is Ponzi account), obtaining 32,752,370 (81.128\%) for $\mathcal{P}^1$ and 9,149 (6.285\%) for $\mathcal{P}^2$ respectively.
It can be seen that the number of metapaths is extremely compressed after merging, but the proportion of interactions involving Ponzi accounts remains high, which supports the assumption of using such predefined metapaths to characterize the interaction behavior of Ponzi accounts.

\subsection{Behavior Refinement and Filtering}
The temporal behavior patterns in Eq.~(\ref{eq: p1p2}) are coarse-grained, because the relationship between the two EOAs before and after, or the relationship between $CA$ and $CA_*$, is undetermined. We further refine the coarse-grained metapaths based on the different behavior patterns between Ponzi and normal accounts, yielding six cases (symbolized as $P^{a,b}$):
\begin{itemize}
  \item $P^\textit{1,1}: EOA_1 \xrightarrow[t_1]{call} CA_\ast \xrightarrow[t_2]{trans} EOA_2$ 
  
  It reflects four behavior patterns: 1) ponzi account accepts new external investment and then rewards old investor; 2) ponzi account accepts new external investment and then transfers commission to the fraudster; 3) fraudster invests and rewards himself to attract external investors; 4) fraudster sweeps commission.

  \item $P^\textit{1,2}: EOA_1 \xrightarrow[t_1]{call} CA_\ast \xrightarrow[t_2]{trans} EOA_1$

  It reflects three behavior patterns: 1) ponzi account accepts new external investment and rewards this investor later; 2) fraudster invests and rewards himself to attract external investors; 3) fraudster sweeps commission.

  \item $P^\textit{2,1}: EOA_1 \xrightarrow[t_1]{call} CA_\ast \xrightarrow[t_2]{call} CA_2 \xrightarrow[t_3]{trans} EOA_2$ 

  This metapath reflects four behavior patterns, which are similar to $P^\textit{1,1}$.

  \item $P^\textit{2,2}: EOA_1 \xrightarrow[t_1]{call} CA_\ast \xrightarrow[t_2]{call} CA_2 \xrightarrow[t_3]{trans} EOA_1$ 

  This metapath reflects three behavior patterns, which are similar to $P^\textit{1,2}$.

  \item 
  $P^\textit{2,3}: EOA_1 \xrightarrow[t_1]{call} CA_\ast \xrightarrow[t_2]{call} CA_\ast \xrightarrow[t_3]{trans} EOA_2$
  
  $P^\textit{2,4}: EOA_1 \xrightarrow[t_1]{call} CA_\ast \xrightarrow[t_2]{call} CA_\ast \xrightarrow[t_3]{trans} EOA_1$

  After analyzing the extracted metapaths, we find that the behavior patterns involving contract self-call only occur in normal interactions.
\end{itemize}
Notably, $a$ indicates the coarse-grained metapath from which this fine-grained metapath is derived. For example, $a=1$ indicates that $P^{1,b}$ is derived from $P^1$. And $b$ is the index of a different fine-grained pattern produced by a further refinement of $P^a$. For example, $P^1$ can further derive two fine-grained metapath patterns, so $b\in\{1,2\}$ when $a=1$; $P^2$ can further derive four fine-grained metapath patterns, so $b\in\{1,2,3,4\}$ when $a=2$.

According to the heterogeneous interaction graph constructed in Table~\ref{tab: graph}, we count the number of all coarse and refined metapaths, as reported in Table~\ref{tab: metapath}.
It can be clearly seen that the number of coarse metapaths is quite large, while the number of different refined metapaths varies significantly. In order to avoid the information redundancy and noise caused by too many metapaths, we introduce an importance-based Top-$K$ filtering strategy to sample the more important metapaths. 
Specifically, we first merge refined symbiotic metapaths into refined super metapaths, during which an importance factor $\omega$, which reflects the number of merged symbiotic metapaths, is derived and appended to the corresponding super metapaths. We believe that the more frequently a class of metapaths occurs, the more helpful it is in characterizing account features by the behavioral patterns it carries.
Then we perform Top-$K$ filtering guided by $\omega$, i.e., sampling the refined super metapaths with importance factors in the top $K$ for subsequent feature aggregation.
As illustrated in Fig.~\ref{fig: method-1}(b), for each kind of refined super metapaths $\mathcal{P}^{1,1}$ (or $\mathcal{P}^{1,2}$), those with larger importance factors will be preserved, such as $\mathcal{P}_1^{1,1}$ (or $\mathcal{P}_2^{1,2}$).

\begin{table}
  \centering
  \caption{Statistics of refined metapaths with and without timestamps.}
  \label{tab: metapath}
  \resizebox{0.8\linewidth}{!}{
  \renewcommand\arraystretch{1.3}
  \begin{tabular}{ll|r|r} 
\hline\hline
\multicolumn{2}{l|}{Metapath} & \textit{w/o} timestamps & \textit{w/} timestamps  \\ 
\hline
\multirow{2}{*}{$P^1$} & $P^{1,1}$    & 6,926,305,286   & 3,275,459,278     \\
                    & $P^{1,2}$       & 9,698,008       & 4,851,608        \\ 
\hline
\multirow{4}{*}{$P^2$} & $P^{2,1}$    & 91,329          & 37,895          \\
                    & $P^{2,2}$       & 3,685           & 2,671           \\
                    & $P^{2,3}$       & 28,191          & 9,229           \\
                    & $P^{2,4}$       & 460             & 254            \\
\hline\hline
\end{tabular}}
\end{table}

\subsection{Metapath Feature Aggregation} \label{sec: feature}
The above two steps aim to obtain the important behavior information represented by super metapaths, and the super metapaths retained during Top-$K$ filtering will participate in the subsequent feature aggregation step.
\emph{TMFAug} is designed to fuse the behavior features represented by metapaths and finally enhance existing Ponzi detection methods on homogeneous Ethereum interaction graph.
However, the preserved super metapaths are still numerous.
In order to alleviate information redundancy and feature explosion during metapath feature aggregation, we adjust the importance factor of super metapaths before doing so.
Specifically, for super metapaths starting from the same head node $v_\textit{eoa}$, we first group them as follows:
\begin{equation}
    \mathcal{M}^\square = \{\mathcal{P}^\square_i \mid \overrightarrow{\mathcal{V}}(\mathcal{P}^\square_i)[1] = v_\textit{eoa}  \} , \quad \text{s.t.} \ \ \square \in \{1,2\}
\end{equation}
where $\overrightarrow{\mathcal{V}}(\mathcal{P})[1]$ represents getting the first element in the node sequence (i.e., head node) of super metapath $\mathcal{P}$.
Then we compute the normalized importance factor for each super metapath $\mathcal{P}^\square_i$ in $\mathcal{M}^\square$ as follows:
\begin{equation}
  \hat{\omega}_\textit{i}^{\square} = \omega_\textit{i}^{\square} / \sum_{\mathcal{P} \in \mathcal{M}^\square}  \omega, \quad \text{s.t.} \ \ \square \in \{1,2\}
\end{equation}

After adjusting the importance factor of all the super metapaths, we then perform feature aggregation to update the CA features.
Specifically, for a target CA node $v_\textit{ca}^\ast$ ($CA_\ast$ in Eq.~(\ref{eq: p1p2})), we first group the super metapaths that contain it as follows:
\begin{equation}
  \mathcal{M}^\square_\ast = \{\mathcal{P}^\square_i \mid \overrightarrow{\mathcal{V}}(\mathcal{P}^\square_i)[2] = v_\textit{ca}^\ast  \} , \quad \text{s.t.} \ \ \square \in \{1,2\}
\end{equation}
After grouping the super metapaths, we update the features of the target CA by aggregating the features of other nodes in the metapath, and the final target CA feature is obtained by processing multiple super metapaths of the same group, as illustrated in Fig.~\ref{fig: method-1}(c).
The process of feature update can be represented as follows:
\begin{equation} 
  \hat{\boldsymbol{x}}_\textit{ca}^\square =  \sum_{\mathcal{P} \in \mathcal{M}^\square_\ast} \hat{\omega} \cdot \sum_{v \in \overrightarrow{\mathcal{V}}(\mathcal{P})} \boldsymbol{x}_v , \quad \text{s.t.} \ \ \square \in \{1,2\}
\end{equation}
where $\boldsymbol{x}_v$ is the initial account feature of nodes along the metapath.
Finally, the updated features $\hat{\boldsymbol{x}}$ contain heterogeneous structural information associated with temporal behavior patterns and will be fed into the downstream classification task.

\section{Experiment} \label{sec:Experiment}
In this section, we evaluate the effectiveness of our \textit{TMFAug} on improving Ponzi scheme detection by answering the following research questions:
\begin{itemize}
  \item \textbf{RQ1}: Can our \emph{TMFAug} improve the performance of Ponzi scheme detection when combined with existing detection methods?
  \item \textbf{RQ2}: Whether time information can help capture more effective behavior patterns?
  \item \textbf{RQ3}: How do different behavior patterns affect the detection results?
  \item \textbf{RQ4}: How does the behavior refinement affect the detection results?
  \item \textbf{RQ5}: How does the behavior filtering affect the detection results?
\end{itemize}

\begin{table*}
  \centering
  \caption{Ponzi detection results of raw methods and their enhanced versions (with \emph{TMFAug} or \emph{MFAug}) in terms of micro-F1 (gain), and gain represents the relative improvement rate. The best enhanced results in each method are marked with boldface. Avg. Rank represents the average rank.}
  \label{tab: result}
  \resizebox{\textwidth}{!}{
    \renewcommand\arraystretch{1.7}
    \begin{tabular}{cc|c|c:c|c:c|c:c} 
\hline\hline
\multicolumn{2}{c|}{Method}                                                    & raw   & \textit{MFAug}($\mathcal{P}^1)$ & \textit{TMFAug}($\mathcal{P}^1)$  & \textit{MFAug}($\mathcal{P}^2)$   & \textit{TMFAug}($\mathcal{P}^2)$  & \textit{MFAug}$(\mathcal{P}^1 + \mathcal{P}^2)$ & \textit{TMFAug}$(\mathcal{P}^1 + \mathcal{P}^2)$  \\ 
\hline
\multirow{2}{*}{\begin{tabular}[c]{@{}l@{}}Manual\\Feature\end{tabular}} & SVM & 73.03 & 71.20 (-2.51\%)        & 71.98 (-1.44\%)          & 73.30 (+0.37\%)          & \textbf{73.83 (+1.11\%)} & 73.03 (+0.00\%)                                 & 72.25 (-1.07\%)                                   \\
                                                                         & RF  & 73.32 & 76.45 (+4.27\%)        & 75.65 (+3.18\%)          & 76.44 (+4.26\%)          & 75.93 (+3.56\%)          & \textbf{76.97 (+4.98\%)}                        & 76.96 (+4.96\%)                                   \\ 
\hline
\multirow{2}{*}{Line}                                                    & SVM & 76.98 & 76.19 (-1.03\%)        & 76.98 (+0.00\%)          & 78.02 (+1.35\%)          & \textbf{78.28 (+1.69\%)} & 75.40 (-2.05\%)                                 & 76.98 (+0.00\%)                                   \\
                                                                         & RF  & 76.97 & 77.49 (+0.68\%)        & 76.18 (-1.03\%)          & 78.27 (+1.69\%)          & \textbf{79.33 (+3.07\%)} & 76.19 (-1.01\%)                                 & 77.23 (+0.32\%)                                   \\ 
\hline
\multirow{2}{*}{Trans2Vec}                                               & SVM & 77.92 & 78.18 (+0.33\%)        & \textbf{80.00 (+2.67\%)} & 77.66 (-0.33\%)          & 78.70 (+1.00\%)          & 78.44 (+0.67\%)                                 & 79.22 (+1.67\%)                                   \\
                                                                         & RF  & 77.92 & 78.44 (+0.66\%)        & 79.22 (+1.67\%)          & 76.88 (-1.33\%)          & 79.48 (+2.00\%)          & 77.92 (+0.00\%)                                 & \textbf{79.96 (+2.62\%)}                          \\ 
\hline
\multirow{2}{*}{DeepWalk}                                                & SVM & 83.00 & 85.09 (+2.52\%)        & 85.09 (+2.52\%)          & 85.09 (+2.52\%)          & 84.82 (+2.19\%)          & 85.09 (+2.52\%)                                 & \textbf{85.09 (+2.52\%)}                          \\
                                                                         & RF  & 81.43 & 83.79 (+2.90\%)        & 84.31 (+3.54\%)          & 83.00 (+1.93\%)          & 84.57 (+3.86\%)          & \textbf{84.83 (+4.18\%)}                        & 83.26 (+2.25\%)                                   \\ 
\hline
\multirow{2}{*}{Node2Vec}                                                & SVM & 84.56 & 86.15 (+1.87\%)        & 86.15 (+1.87\%)          & 85.08 (+0.61\%)          & 85.36 (+0.95\%)          & \textbf{86.67 (+2.50\%)}                        & 86.15 (+1.87\%)                                   \\
                                                                         & RF  & 86.13 & 86.40 (+0.31\%)        & 86.92 (+0.92\%)          & \textbf{87.19 (+1.23\%)} & 86.41 (+0.33\%)          & 86.66 (+0.62\%)                                 & 86.66 (+0.62\%)                                   \\ 
\hline
\multicolumn{2}{c|}{LightGBM~\cite{zhang2021detecting}}                        & 75.14 & 76.98 (+2.45\%)        & \textbf{79.07 (+5.23\%)} & 77.23 (+2.78\%)           & 78.54 (+4.52\%)           & 75.40 (+0.35\%)                                 & 77.25 (+2.81\%)                                    \\ 
\hline      
\multicolumn{2}{c|}{GCN}                                                       & 84.50 & 85.71 (+1.43\%)        & 85.71 (+1.43\%)          & 85.71 (+1.43\%)          & \textbf{86.39 (+2.24\%)} & 85.60 (+1.30\%)                                 & 85.92 (+1.68\%)                                   \\ 
\hline      
\multicolumn{2}{c|}{GAT}                                                       & 84.24 & 87.06 (+3.35\%)        & 87.59 (+3.98\%)          & 86.60 (+2.80\%)          & 87.12 (+3.42\%)          & 87.06 (+3.35\%)                                 & \textbf{87.68 (+4.08\%)}                          \\ 
\hline      
\multicolumn{2}{c|}{GIN}                                                       & 84.59 & 85.33 (+0.87\%)        & 85.08 (+0.58\%)          & 85.86 (+1.50\%)          & \textbf{86.38 (+2.12\%)} & 86.12 (+1.81\%)                                 & 85.08 (+0.58\%)                                   \\ 
\hline      
\multicolumn{2}{c|}{SAGE}                                                      & 85.70 & 86.75 (+1.23\%)        & \textbf{86.91 (+1.41\%)} & 86.65 (+1.11\%)          & 86.86 (+1.35\%)          & 86.06 (+0.42\%)                                 & 86.59 (+1.04\%)                                   \\ 
\hline      
\multicolumn{2}{c|}{FAGNN}                                                     & 85.34 & 85.92 (+0.68\%)        & 86.07 (+0.86\%)          & 85.76 (+0.49\%)          & \textbf{86.55 (+1.42\%)} & 85.76 (+0.49\%)                                 & 85.92 (+0.68\%)                                   \\ 
\hline      
\multicolumn{2}{c|}{Avg. Rank}                                                  & 6.19  & 3.94                   & 3.00                     & 3.94                     & 2.56                     & 3.81                                            & 2.94                                              \\
\hline\hline
\end{tabular}}
   \end{table*}

\subsection{Ponzi Detection Methods and Experimental Setup}
To illustrate the effectiveness and generality of our \emph{TMFAug} module, we combine it with three categories of Ponzi detection methods: manual feature engineering, graph embedding and GNN-based methods.
To illustrate the superiority of our \emph{TMFAug} module, we introduce a \emph{MFAug} module and the \emph{HFAug} module~\cite{HFAug} for comparison. The differences between the three modules are shown in the Table~\ref{tab: HFAug-MFAug-TMFAug}.

Manual feature engineering is the most common approach for Ponzi detection, we use 15 manual features listed in Sec.~\ref{sec: manual-feature} to construct the initial features for account nodes.
For Ponzi detection based on graph embedding, we consider Line, DeepWalk, Node2Vec and Trans2Vec algorithms for constructing initial account node features, respectively.
For the above two categories of methods, we achieve Ponzi detection by feeding the generated account features into two machine learning classifiers: Support Vector Machine (SVM)~\cite{noble2006support} and Random Forest (RF)~\cite{segal2004machine}. Besides, we use LightGBM from Zhang et al.~\cite{zhang2021detecting} as a new classifier for comparison experiments.
When combined with \emph{TMFAug}, we feed the updated features generated via \emph{TMFAug} into downstream classifiers.
For GNN-based methods, we use four popular GNNs: GCN, GAT, GIN, SAGE and FAGNN, with the same initial features as manual feature engineering.

For DeepWalk and Node2Vec, we set the dimension of embedding, window size, walk length and the number of walks per node to 128, 10, 50 and 5 respectively. For Node2Vec, we perform a grid search of return parameter $p$ and in-out parameter $q$ in $\{0.5, 1, 2\}$. 
For Trans2Vec, we set the dimension of embedding, window size, walk length and the number of walks per node to 64, 10, 50, 5 respectively.
For Line, we set the embedding dimension and order to 32 and 2, respectively. 
For GNN-based methods, we set the hidden dimension of GCN, GAT, GIN, SAGE and FAGNN to 128, 32, 128, 128 and 128 respectively, and the learning rate to 0.01, 0.1, 0.1, 0.01 and 0.01 respectively.
For filtering parameter $K$, we vary in \{1\%, 3\%, 5\%, 7\%, 9\%, 10\%, 20\%, 30\%, 40\%, 50\%\} and choose the best results for Ponzi detection.
For all methods, we repeat 5-fold cross-validation five times with five different random seeds and report the average micro-F1 score over $5\times 5=25$ experiments.

\subsection{Enhancement for Ponzi Dectection (\textbf{RQ1})}

Table~\ref{tab: result} reports the results of performance comparison between the raw methods and their enhanced version (with \emph{TMFAug}), from which we can observe that there is a significant boost in detection performance across all methods. 
Overall, these detection methods combined with \emph{TMFAug} obtain higher average detection performance in most cases, and the \emph{TMFAug} achieves a 93.75\% success rate\footnote{The success rate refers to the percentage of enhanced methods with F1 score higher than that of the corresponding raw methods.} on improving Ponzi detection.
Specifically, except for the partial results of manual feature engineering and Line, our \emph{TMFAug} consistently enhances detection performance, yielding 1.11\% $\sim$ 5.23\%, 0.32\% $\sim$ 3.86\%, 0.58\% $\sim$ 4.08\% relative improvement on the three categories of detection methods, respectively.
It is noteworthy that detection methods exhibiting higher initial performance are generally more likely to experience positive gains from our \emph{TMFAug}.
Meanwhile, our \textit{TMFAug} achieves a 100\% success rate when combining with state-of-the-art detection approaches, suggesting its powerful generality.
In this regard, we make the following reasonable explanation.
Both random-walk-based methods and GNN-based methods are capable of acquiring structured account features during detection, while our module can also capture structured behavior features. The combination of these two aspects brings consistent positive gains.

\begin{figure}
    \centering
    \includegraphics[width=\linewidth]{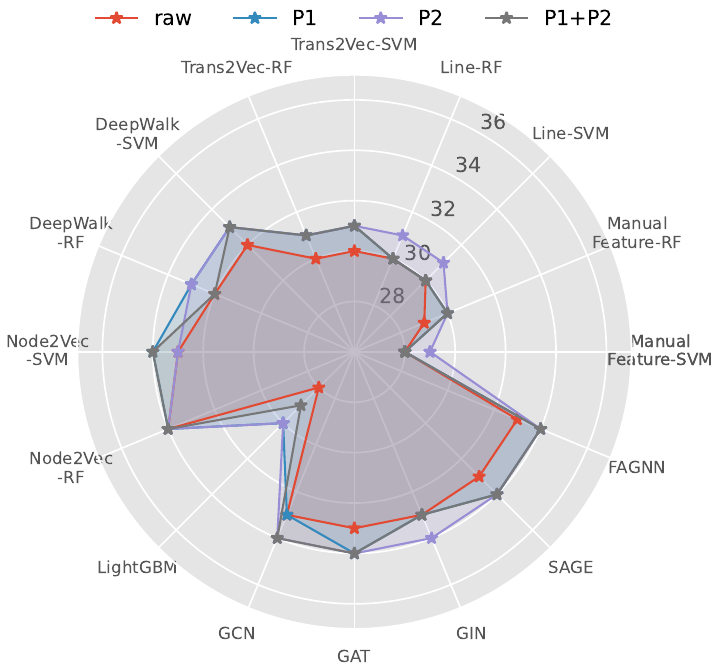}
    \caption{The difference in the number of detected Ponzi accounts before and after applying \emph{TMFAug}.}
    \label{fig:radar}
\end{figure}

Fig.~\ref{fig:radar} shows the difference in the number of Ponzi accounts detected by different Ponzi detection algorithms before and after applying \emph{TMFAug}, from which we can observe that our module effectively helps various Ponzi detection algorithms to identify more Ponzi accounts, demonstrating its effectiveness and generality. However, it should be noted that due to the limited number of Ponzi account samples in the testing set (only 38), the increase in identified Ponzi accounts after applying \emph{TMFAug} may be relatively small.

\begin{figure}
  \centering
  \includegraphics[width=\linewidth]{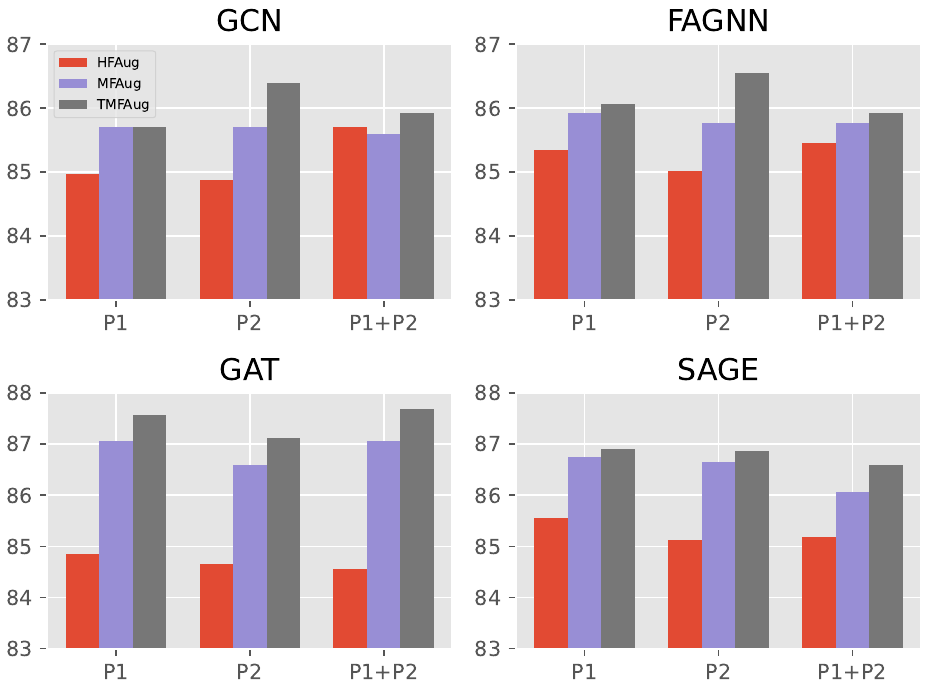}
  \caption{The results of \textit{HFAug}, \textit{MFAug}, and \textit{TMFAug} in Ponzi detection.}
  \label{fig: HvsT}
\end{figure}

Fig.~\ref{fig: HvsT} illustrates the performance difference between the three different feature augmentation modules on GNN-based detection algorithms, from which it can be seen that our \emph{TMFAug} consistently achieves the best results across all settings, validating its superiority. Specifically, comparing \emph{MFAug} and \emph{HFAug}, the former's performance advantage stems from utilizing a large number of refined metapaths to capture diverse interaction features. Comparing \emph{MFAug} and \emph{TMFAug}, the performance advantage of the latter benefits from utilizing time-aware metapaths to capture dynamic interaction features.

These phenomena provide a positive answer to \textbf{RQ1}, indicating that the \emph{TMFAug} module can benefit the existing Ponzi detection methods via feature augmentation and improve their performance without adjusting them.

\begin{table}
  \centering
  \caption{Summary of the differences in different feature augmentation modules.}
  \label{tab: HFAug-MFAug-TMFAug}
    \resizebox{\linewidth}{!}{
      \renewcommand\arraystretch{1.2}
  \begin{tabular}{lccc} 
  \hline\hline
  \multirow{2}{*}{Module} & \multicolumn{3}{c}{Design}                       \\ 
  \cline{2-4}
                          & Timestamp    & Refinement   & Num. of Metapaths  \\ 
  \hline
  \textit{HFAug}          &              &              & One                \\
  \textit{MFAug}          &              & $\checkmark$ & Multiple           \\ 
  \textit{TMFAug}         & $\checkmark$ & $\checkmark$ & Multiple           \\
  \hline\hline
  \end{tabular}}
  \end{table}

\subsection{Superiority of Temporal Information (\textbf{RQ2})}
After evaluating the overall performance of our method, we further investigate the superiority of time-aware metapaths over time-less ones.
By comparing the two modules (\emph{TMFAug} and \emph{MFAug}) in Table~\ref{tab: result}, 
we observe that methods with \emph{TMFAug} outperform those with \emph{MFAug} at most cases.
Specifically, \emph{TMFAug} achieves a higher average performance ranking than \emph{MFAug}, i.e., $Rank(3.00) > Rank(3.94)$ in $\mathcal{P}^1$, $Rank(2.56) > Rank(3.94)$ in $\mathcal{P}^2$, $Rank(2.94) > Rank(3.81)$ in $(\mathcal{P}^1 + \mathcal{P}^2)$.
Taking Trans2Vec as an example, in the experiment of integrating $\mathcal{P}^2$ metapaths, \textit{MFAug} that neglect dynamics brings negative gains for identifying Ponzi accounts.
These phenomena provide a positive answer to \textbf{RQ2}, indicating that the temporal behavior patterns captured by the \emph{TMFAug} module are more effective for improving Ponzi detection than timeless ones.

\subsection{Impact of Behavior Pattern (\textbf{RQ3})}
We further investigate the impact of different behavior patterns on detecting Ponzi schemes.
Specifically, we compare a total of six combinations of different metapaths ($\mathcal{P}^1$, $\mathcal{P}^2$ and $\mathcal{P}^1 + \mathcal{P}^2$) and different modules (\emph{TMFAug} and \emph{MFAug}), as shown in Table~\ref{tab: result}.
As we can see, for \emph{TMFAug} or \emph{MFAug}, the performance ranking of different metapaths is consistent: $\textit{TMFAug}(\mathcal{P}^2) > \textit{TMFAug}(\mathcal{P}^1 + \mathcal{P}^2) > \textit{TMFAug}(\mathcal{P}^1)$ and $\textit{MFAug}(\mathcal{P}^2) > \textit{MFAug}(\mathcal{P}^1 + \mathcal{P}^2) > \textit{MFAug}(\mathcal{P}^1)$, which suggests that the enhancement effect relies on the choice of metapaths and answers to \textbf{RQ3}.
Both $\mathcal{P}^1$ and $\mathcal{P}^2$ are extracted from the basic behavior patterns of Ponzi and normal accounts defined in Eq.~(\ref{eq: p1p2}), and we have reasonable explanations for their performance difference.
As shown in Table~\ref{tab: metapath}, the number of $P^1$ far exceeds that of $P^2$, even with the existence of relation merging and behavior filtering, the number of retained super metapaths is still huge, which inevitably leads to information redundancy and over-smoothing during feature aggregation.
On the other hand, compared with $P^1$, $P^2$ contains more diverse and fine-grained behavior patterns, which can better characterize the differences between Ponzi accounts and normal accounts.
In addition, this is also illustrated by the results on manual feature engineering and Line, e.g., aggregating features from $\mathcal{P}^1$ may lead to a negative gain in performance.

\begin{table*}
  \centering
  \caption{Ablation study on behavior refinement and filtering. (w/o refinement) represents the \emph{TMFAug} without behavior refinement, and (w/o filtering) represents the \textit{TMFAug} without behavior filtering. The bold marker represents the better result between \emph{TMFAug} with and without behavior refinement, and the underline represents the better result between \emph{TMFAug} with and without behavior filtering. Avg.local rank represents the average rank with different super metapath.}
  \label{tab: AS}
  \resizebox{\linewidth}{!}{
  \renewcommand\arraystretch{1.4}
  \begin{tabular}{l|cc|cc|cc|cc|cc|c|c|c|c|c|c|c} 
\hline\hline
\multirow{2}{*}{Method}                                           & \multicolumn{2}{c|}{Manual Feature}             & \multicolumn{2}{c|}{Line}                       & \multicolumn{2}{c|}{Trans2Vec}                  & \multicolumn{2}{c|}{DeepWalk}                   & \multicolumn{2}{c|}{Node2Vec}           & \multirow{2}{*}{\begin{tabular}[c]{@{}c@{}}LightGBM\end{tabular}} & \multirow{2}{*}{GCN}   & \multirow{2}{*}{GAT}   & \multirow{2}{*}{GIN}   & \multirow{2}{*}{SAGE}  & \multirow{2}{*}{FAGNN} & \multirow{2}{*}{\begin{tabular}[c]{@{}c@{}}Avg.~ \\Local Rank\end{tabular}}  \\ 
\cline{2-11}
                                                                  & SVM                    & RF                     & SVM                    & RF                     & SVM                    & RF                     & SVM                    & RF                     & SVM                    & RF             &                           &                        &                        &                        &                        &                        &                                                                              \\ 
\hline
\textit{TMFAug}$(\mathcal{P}^1)$ (w/o refinement)                 & \textbf{72.26}         & \textbf{76.19}         & 76.98                  & 75.91                  & 78.18                  & 77.92                  & 85.09                  & 84.05                  & 86.14                  & \textbf{87.18} & 76.70                     & \textbf{85.76}         & 87.32                  & \textbf{85.34}         & 86.62                  & 85.39                  & 1.88                                                                         \\
\textit{TMFAug}$(\mathcal{P}^1)$ (w/o filtering)                  & 67.28                  & 70.17                  & 58.13                  & 68.63                  & 77.66                  & \uline{79.74}          & 82.47                  & \uline{84.57}          & 86.13                  & 85.88          & 69.65                     & \uline{86.43}          & 83.25                  & \uline{86.64}          & 84.97                  & 86.07                  & 2.38                                                                         \\
\textit{TMFAug}$(\mathcal{P}^1)$                                  & \uline{71.98}          & \uline{75.65}          & \uline{76.98}          & \textbf{\uline{76.18}} & \textbf{\uline{80.00}} & \textbf{79.22}         & \uline{85.09}          & \textbf{84.31}         & \textbf{\uline{86.15}} & \uline{86.92}  & \textbf{\uline{79.07}}    & 85.71                  & \textbf{\uline{87.59}} & 85.08                  & \textbf{\uline{86.91}} & \textbf{\uline{86.07}} & \textbf{1.56}                                                                \\ 
\hline
\textit{TMFAug}$(\mathcal{P}^2)$ (w/o refinement)                 & 73.82                  & 75.67                  & 77.76                  & 79.32                  & 76.88                  & 77.40                  & \textbf{85.09}         & 83.79                  & 85.36                  & \textbf{86.93} & 77.22                     & 86.18                  & \textbf{87.22}         & 85.86                  & 86.65                  & 85.55                  & 2.06                                                                         \\
\textit{TMFAug}$(\mathcal{P}^2)$ (w/o filtering)                  & 71.72                  & 74.09                  & 77.74                  & 78.26                  & 77.66                  & 78.18                  & 84.06                  & 82.47                  & 84.30                  & \uline{86.92}  & 75.39                     & \uline{86.55}          & 86.38                  & 84.55                  & 86.85                  & 86.13                  & 2.56                                                                         \\
\textit{TMFAug}$(\mathcal{P}^2)$                                  & \textbf{\uline{73.83}} & \textbf{\uline{75.93}} & \textbf{\uline{78.28}} & \textbf{\uline{79.33}} & \textbf{\uline{78.70}} & \textbf{\uline{79.48}} & \uline{84.82}          & \textbf{\uline{84.57}} & \uline{85.36}          & 86.41          & \textbf{\uline{78.54}}    & \textbf{86.39}         & \uline{87.12}          & \textbf{\uline{86.38}} & \textbf{\uline{86.86}} & \textbf{\uline{86.55}} & \textbf{1.31}                                                                \\ 
\hline
\textit{TMFAug}$(\mathcal{P}^1 + \mathcal{P}^2)$ (w/o refinement) & \textbf{72.51}         & 75.66                  & \textbf{78.29}         & 77.23                  & 77.14                  & 75.32                  & 85.09                  & \textbf{84.57}         & 85.63                  & 86.66          & 76.98                     & 85.60                  & \textbf{87.74}         & 85.07                  & \textbf{86.96}         & 85.34                  & 1.88                                                                         \\
\textit{TMFAug}$(\mathcal{P}^1 + \mathcal{P}^2)$ (w/o filtering)  & 67.27                  & 71.22                  & 57.09                  & 66.76                  & \uline{80.78}          & 78.44                  & 84.05                  & \uline{83.52}          & 85.87                  & \uline{86.92}  & 70.95                     & 85.91                  & 84.61                  & \uline{87.44}          & 83.13                  & 85.34                  & 2.31                                                                         \\
\textit{TMFAug}$(\mathcal{P}^1 + \mathcal{P}^2)$                  & \uline{72.25}          & \textbf{\uline{76.96}} & \uline{76.98}          & \uline{77.23}          & \textbf{79.22}         & \textbf{\uline{79.96}} & \textbf{\uline{85.09}} & 83.26                  & \textbf{\uline{86.15}} & \textbf{86.66} & \textbf{\uline{77.25}}    & \textbf{\uline{85.92}} & \uline{87.68}          & \textbf{85.08}         & \uline{86.59}          & \textbf{\uline{85.92}} & \textbf{1.56}                                                                \\
\hline\hline
\end{tabular}}
\end{table*}

\subsection{Impact of Behavior Refinement (\textbf{RQ4})}
Behavior refinement is proposed to characterize different behavior patterns in fine granularity, and we further investigate its effectiveness.
Table~\ref{tab: AS} reports the performance comparison of \emph{TMFAug} with and without behavior refinement.
Overall, we can observe that \emph{TMFAug} with behavior refinement achieves a higher average local ranking across different metapaths, i.e., $Rank(1.56) > Rank(1.88)$ in $\mathcal{P}^1$, $Rank(1.31) > Rank(2.06)$ in $\mathcal{P}^2$, and $Rank(1.56) > Rank(1.88)$ in $(\mathcal{P}^1 + \mathcal{P}^2)$.
Furthermore, behavior refinement works better on $\mathcal{P}^2$, manifested as a larger ranking boost.
Here we combine the properties of different metapaths to provide further analysis.
First, $\mathcal{P}^1$ has a straightforward structure, so we can just refine it into two cases.
Despite the behavior refinement, simple interactions do not lead to more discriminative patterns, nor do they further yield gains.
While $\mathcal{P}^2$ has a more complex structure and contains more behavior patterns, especially patterns unique to normal accounts, and thus can be refined into four cases.
Behavior refinement can process coarse-grained information into fine-grained and discriminative patterns, so it is more beneficial for more complex $\mathcal{P}^2$.
Moreover, behavior refinement can help to better perform subsequent behavior filtering.
These phenomena provide a positive answer to \textbf{RQ4}, indicating that the \textit{TMFAug} somewhat relies on behavior refinement, and the more complex interaction patterns access to their superior performance.

\subsection{Impact of Behavior Filtering (\textbf{RQ5})}
Behavior filtering is proposed to alleviate information redundancy and feature explosion during metapath feature aggregation, and we further investigate its effectiveness.
Table~\ref{tab: AS} reports the performance comparison of \emph{TMFAug} with and without behavior filtering.
Overall, we can observe that \emph{TMFAug} with behavior filtering achieves a higher average local ranking across different metapaths, i.e., $Rank(1.56) > Rank(2.38)$ in $\mathcal{P}^1$, $Rank(1.31) > Rank(2.56)$ in $\mathcal{P}^2$, and $Rank(1.56) > Rank(2.31)$ in $(\mathcal{P}^1 + \mathcal{P}^2)$, indicating its effectiveness.
In this regard, we make the following reasonable explanation.
Without behavior filtering, metapaths are not only redundant in number, but also contain much unimportant information.
When updating node features along metapaths, using too many metapaths could make the information jumbled, and those unimportant metapaths will interfere with the subsequent detection.
In addition, behavior filtering is more beneficial for manual feature engineering and Line, in which the highest relative improvement rate reaches 34.84\%. 
We speculate that integrating excessive metapaths will greatly destroy the original feature representation of manual feature engineering and Line due to the weak robustness of their initial features.
And for Trans2Vec, DeepWalk, Node2Vec and GNN-based methods, behavior filtering also yields some gains, but only marginally.
The reason for the low gain is that the learned embeddings are robust and have richer structural and semantic information than the manual feature engineering and Line, so noisy metapaths do not severely affect them.
These phenomena provide a positive answer to \textbf{RQ5}, indicating that the behavior filtering can effectively alleviate information redundancy and feature explosion during metapath feature aggregation, further improving Ponzi detection.

\section{Conclusion} \label{sec:conclusion}
Existing methods for Ponzi scheme detection usually neglect complex and temporal interaction behaviors.
In this paper, we propose a Time-aware Metapath Feature Augmentation module, which includes symbiotic relationship merging, behavior refinement and filtering, and metapath feature aggregation.
This module can effectively capture the real metapath-based transaction patterns and aggregate the temporal behavior information during Ponzi scheme detection on Ethereum.
Experiments show that this module can significantly improve the performance of exciting Ponzi scheme detection methods. 
Light of the fact that the current method requires expert knowledge to capture metapaths, which is time-consuming. 
Our future research will attempt to reduce the manual definition of metapaths, for instance by automatically learning metapaths~\cite{wang2021embedding,chang2022megnn,yun2019graph} to save time and improve generalization.

\bibliographystyle{IEEEtran}
\bibliography{TMFAug}

\begin{thebibliography}{10}
\providecommand{\url}[1]{#1}
\csname url@samestyle\endcsname
\providecommand{\newblock}{\relax}
\providecommand{\bibinfo}[2]{#2}
\providecommand{\BIBentrySTDinterwordspacing}{\spaceskip=0pt\relax}
\providecommand{\BIBentryALTinterwordstretchfactor}{4}
\providecommand{\BIBentryALTinterwordspacing}{\spaceskip=\fontdimen2\font plus
\BIBentryALTinterwordstretchfactor\fontdimen3\font minus \fontdimen4\font\relax}
\providecommand{\BIBforeignlanguage}[2]{{%
\expandafter\ifx\csname l@#1\endcsname\relax
\typeout{** WARNING: IEEEtran.bst: No hyphenation pattern has been}%
\typeout{** loaded for the language `#1'. Using the pattern for}%
\typeout{** the default language instead.}%
\else
\language=\csname l@#1\endcsname
\fi
#2}}
\providecommand{\BIBdecl}{\relax}
\BIBdecl

\bibitem{li2020survey}
X.~Li, P.~Jiang, T.~Chen, X.~Luo, and Q.~Wen, ``A survey on the security of blockchain systems,'' \emph{Future Generation Computer Systems}, vol. 107, pp. 841--853, 2020.

\bibitem{stinson2005cryptography}
D.~R. Stinson, \emph{Cryptography: theory and practice}.\hskip 1em plus 0.5em minus 0.4em\relax Chapman and Hall/CRC, 2005.

\bibitem{lashkari2021comprehensive}
B.~Lashkari and P.~Musilek, ``A comprehensive review of blockchain consensus mechanisms,'' \emph{IEEE Access}, vol.~9, pp. 43\,620--43\,652, 2021.

\bibitem{nakamoto2008bitcoin}
S.~Nakamoto, ``Bitcoin: A peer-to-peer electronic cash system,'' \emph{Decentralized Business Review}, p. 21260, 2008.

\bibitem{buterin2014next}
V.~Buterin \emph{et~al.}, ``A next-generation smart contract and decentralized application platform,'' \emph{white paper}, vol.~3, no.~37, pp. 2--1, 2014.

\bibitem{wood2014ethereum}
G.~Wood \emph{et~al.}, ``Ethereum: A secure decentralised generalised transaction ledger,'' \emph{Ethereum project yellow paper}, vol. 151, no. 2014, pp. 1--32, 2014.

\bibitem{wang2019blockchain}
S.~Wang, L.~Ouyang, Y.~Yuan, X.~Ni, X.~Han, and F.-Y. Wang, ``Blockchain-enabled smart contracts: architecture, applications, and future trends,'' \emph{IEEE Transactions on Systems, Man, and Cybernetics: Systems}, vol.~49, no.~11, pp. 2266--2277, 2019.

\bibitem{artzrouni2009mathematics}
M.~Artzrouni, ``The mathematics of ponzi schemes,'' \emph{Mathematical Social Sciences}, vol.~58, no.~2, pp. 190--201, 2009.

\bibitem{kell2021forsage}
T.~Kell, H.~Yousaf, S.~Allen, S.~Meiklejohn, and A.~Juels, ``Forsage: Anatomy of a smart-contract pyramid scheme,'' \emph{arXiv preprint arXiv:2105.04380}, 2021.

\bibitem{macpherson2011survey}
D.~Macpherson, ``A survey of homogeneous structures,'' \emph{Discrete Mathematics}, vol. 311, no.~15, pp. 1599--1634, 2011.

\bibitem{Ethident}
J.~Zhou, C.~Hu, J.~Chi, J.~Wu, M.~Shen, and Q.~Xuan, ``Behavior-aware account de-anonymization on ethereum interaction graph,'' \emph{IEEE Transactions on Information Forensics and Security}, vol.~17, pp. 3433--3448, 2022.

\bibitem{jin2023ethereum}
J.~Jin, J.~Zhou, W.~Chen, Y.~Sheng, and Q.~Xuan, ``Ethereum’s ponzi scheme detection work based on graph ideas,'' in \emph{Deep Learning Applications: In Computer Vision, Signals and Networks}.\hskip 1em plus 0.5em minus 0.4em\relax World Scientific, 2023, pp. 215--241.

\bibitem{shi2016survey}
C.~Shi, Y.~Li, J.~Zhang, Y.~Sun, and S.~Y. Philip, ``A survey of heterogeneous information network analysis,'' \emph{IEEE Transactions on Knowledge and Data Engineering}, vol.~29, no.~1, pp. 17--37, 2016.

\bibitem{sun2011pathsim}
Y.~Sun, J.~Han, X.~Yan, P.~S. Yu, and T.~Wu, ``Pathsim: Meta path-based top-k similarity search in heterogeneous information networks,'' \emph{Proceedings of the VLDB Endowment}, vol.~4, no.~11, pp. 992--1003, 2011.

\bibitem{dong2017metapath2vec}
Y.~Dong, N.~V. Chawla, and A.~Swami, ``{Metapath2vec: Scalable Representation Learning for Heterogeneous Networks},'' in \emph{Proceedings of the 23rd ACM SIGKDD international conference on knowledge discovery and data mining}, 2017, pp. 135--144.

\bibitem{shi2018heterogeneous}
C.~Shi, B.~Hu, W.~X. Zhao, and S.~Y. Philip, ``Heterogeneous information network embedding for recommendation,'' \emph{IEEE Transactions on Knowledge and Data Engineering}, vol.~31, no.~2, pp. 357--370, 2018.

\bibitem{gong2023neighborhood}
S.~Gong, J.~Zhou, C.~Xie, and Q.~Xuan, ``Neighborhood homophily-based graph convolutional network,'' in \emph{Proceedings of the 32nd ACM International Conference on Information and Knowledge Management}, 2023, pp. 3908--3912.

\bibitem{gong2023clarify}
S.~Gong, J.~Zhou, and Q.~Xuan, ``Clarify confused nodes through separated learning,'' \emph{arXiv preprint arXiv:2306.02285}, 2023.

\bibitem{fu2018link}
C.~Fu, M.~Zhao, L.~Fan, X.~Chen, J.~Chen, Z.~Wu, Y.~Xia, and Q.~Xuan, ``Link weight prediction using supervised learning methods and its application to yelp layered network,'' \emph{IEEE Transactions on Knowledge and Data Engineering}, vol.~30, no.~8, pp. 1507--1518, 2018.

\bibitem{yu2019target}
S.~Yu, M.~Zhao, C.~Fu, J.~Zheng, H.~Huang, X.~Shu, Q.~Xuan, and G.~Chen, ``Target defense against link-prediction-based attacks via evolutionary perturbations,'' \emph{IEEE Transactions on Knowledge and Data Engineering}, vol.~33, no.~2, pp. 754--767, 2019.

\bibitem{zhou2021robustecd}
J.~Zhou, Z.~Chen, M.~Du, L.~Chen, S.~Yu, G.~Chen, and Q.~Xuan, ``Robustecd: Enhancement of network structure for robust community detection,'' \emph{IEEE Transactions on Knowledge and Data Engineering}, vol.~35, no.~1, pp. 842--856, 2021.

\bibitem{su2022comprehensive}
X.~Su, S.~Xue, F.~Liu, J.~Wu, J.~Yang, C.~Zhou, W.~Hu, C.~Paris, S.~Nepal, D.~Jin \emph{et~al.}, ``A comprehensive survey on community detection with deep learning,'' \emph{IEEE Transactions on Neural Networks and Learning Systems}, 2022.

\bibitem{xuan2019subgraph}
Q.~Xuan, J.~Wang, M.~Zhao, J.~Yuan, C.~Fu, Z.~Ruan, and G.~Chen, ``Subgraph networks with application to structural feature space expansion,'' \emph{IEEE Transactions on Knowledge and Data Engineering}, vol.~33, no.~6, pp. 2776--2789, 2019.

\bibitem{zhou2020m}
J.~Zhou, J.~Shen, S.~Yu, G.~Chen, and Q.~Xuan, ``M-evolve: structural-mapping-based data augmentation for graph classification,'' \emph{IEEE Transactions on Network Science and Engineering}, vol.~8, no.~1, pp. 190--200, 2020.

\bibitem{perozzi2014deepwalk}
B.~Perozzi, R.~Al-Rfou, and S.~Skiena, ``Deepwalk: Online learning of social representations,'' in \emph{Proceedings of the 20th ACM SIGKDD international conference on Knowledge discovery and data mining}, 2014, pp. 701--710.

\bibitem{grover2016node2vec}
A.~Grover and J.~Leskovec, ``node2vec: Scalable feature learning for networks,'' in \emph{Proceedings of the 22nd ACM SIGKDD international conference on Knowledge discovery and data mining}, 2016, pp. 855--864.

\bibitem{tang2015line}
J.~Tang, M.~Qu, M.~Wang, M.~Zhang, J.~Yan, and Q.~Mei, ``Line: Large-scale information network embedding,'' in \emph{Proceedings of the 24th international conference on world wide web}, 2015, pp. 1067--1077.

\bibitem{kipf2016semi}
T.~N. Kipf and M.~Welling, ``Semi-supervised classification with graph convolutional networks,'' \emph{arXiv preprint arXiv:1609.02907}, 2016.

\bibitem{velickovic2017graph}
P.~Velickovic, G.~Cucurull, A.~Casanova, A.~Romero, P.~Lio, and Y.~Bengio, ``Graph attention networks,'' \emph{stat}, vol. 1050, p.~20, 2017.

\bibitem{hamilton2017inductive}
W.~Hamilton, Z.~Ying, and J.~Leskovec, ``Inductive representation learning on large graphs,'' \emph{Advances in neural information processing systems}, vol.~30, 2017.

\bibitem{xu2018powerful}
K.~Xu, W.~Hu, J.~Leskovec, and S.~Jegelka, ``How powerful are graph neural networks?'' \emph{arXiv preprint arXiv:1810.00826}, 2018.

\bibitem{moore2012postmodern}
T.~Moore, J.~Han, and R.~Clayton, ``The postmodern ponzi scheme: Empirical analysis of high-yield investment programs,'' in \emph{International Conference on financial cryptography and data security}.\hskip 1em plus 0.5em minus 0.4em\relax Springer, 2012, pp. 41--56.

\bibitem{vasek2015there}
M.~Vasek and T.~Moore, ``There’s no free lunch, even using bitcoin: Tracking the popularity and profits of virtual currency scams,'' in \emph{International conference on financial cryptography and data security}.\hskip 1em plus 0.5em minus 0.4em\relax Springer, 2015, pp. 44--61.

\bibitem{chen2018detecting}
W.~Chen, Z.~Zheng, J.~Cui, E.~Ngai, P.~Zheng, and Y.~Zhou, ``Detecting ponzi schemes on ethereum: Towards healthier blockchain technology,'' in \emph{Proceedings of the 2018 world wide web conference}, 2018, pp. 1409--1418.

\bibitem{fan2020expose}
S.~Fan, S.~Fu, H.~Xu, and C.~Zhu, ``Expose your mask: smart ponzi schemes detection on blockchain,'' in \emph{2020 International Joint Conference on Neural Networks (IJCNN)}.\hskip 1em plus 0.5em minus 0.4em\relax IEEE, 2020, pp. 1--7.

\bibitem{bartoletti2020dissecting}
M.~Bartoletti, S.~Carta, T.~Cimoli, and R.~Saia, ``Dissecting ponzi schemes on ethereum: identification, analysis, and impact,'' \emph{Future Generation Computer Systems}, vol. 102, pp. 259--277, 2020.

\bibitem{chen2021sadponzi}
W.~Chen, X.~Li, Y.~Sui, N.~He, H.~Wang, L.~Wu, and X.~Luo, ``Sadponzi: Detecting and characterizing ponzi schemes in ethereum smart contracts,'' \emph{Proceedings of the ACM on Measurement and Analysis of Computing Systems}, vol.~5, no.~2, pp. 1--30, 2021.

\bibitem{krupp2018teether}
J.~Krupp and C.~Rossow, ``$\{$teEther$\}$: Gnawing at ethereum to automatically exploit smart contracts,'' in \emph{27th USENIX Security Symposium (USENIX Security 18)}, 2018, pp. 1317--1333.

\bibitem{chen2021improving}
Y.~Chen, H.~Dai, X.~Yu, W.~Hu, Z.~Xie, and C.~Tan, ``Improving ponzi scheme contract detection using multi-channel textcnn and transformer,'' \emph{Sensors}, vol.~21, no.~19, p. 6417, 2021.

\bibitem{zhang2021detecting}
Y.~Zhang, W.~Yu, Z.~Li, S.~Raza, and H.~Cao, ``Detecting ethereum ponzi schemes based on improved lightgbm algorithm,'' \emph{IEEE Transactions on Computational Social Systems}, 2021.

\bibitem{yu2021ponzi}
S.~Yu, J.~Jin, Y.~Xie, J.~Shen, and Q.~Xuan, ``Ponzi scheme detection in ethereum transaction network,'' in \emph{International Conference on Blockchain and Trustworthy Systems}.\hskip 1em plus 0.5em minus 0.4em\relax Springer, 2021, pp. 175--186.

\bibitem{liu2022fa}
J.~Liu, J.~Zheng, J.~Wu, and Z.~Zheng, ``Fa-gnn: Filter and augment graph neural networks for account classification in ethereum,'' \emph{IEEE Transactions on Network Science and Engineering}, vol.~9, no.~4, pp. 2579--2588, 2022.

\bibitem{wu2020phishers}
J.~Wu, Q.~Yuan, D.~Lin, W.~You, W.~Chen, C.~Chen, and Z.~Zheng, ``Who are the phishers? phishing scam detection on ethereum via network embedding,'' \emph{IEEE Transactions on Systems, Man, and Cybernetics: Systems}, vol.~52, no.~2, pp. 1156--1166, 2020.

\bibitem{jin2022dual}
J.~Jin, J.~Zhou, C.~Jin, S.~Yu, Z.~Zheng, and Q.~Xuan, ``Dual-channel early warning framework for ethereum ponzi schemes,'' \emph{The 7th China National Conference on Big Data and Social Computing}, 2022.

\bibitem{HFAug}
C.~Jin, J.~Jin, J.~Zhou, J.~Wu, and Q.~Xuan, ``Heterogeneous feature augmentation for ponzi detection in ethereum,'' \emph{IEEE Transactions on Circuits and Systems II: Express Briefs}, 2022.

\bibitem{noble2006support}
W.~S. Noble, ``What is a support vector machine?'' \emph{Nature biotechnology}, vol.~24, no.~12, pp. 1565--1567, 2006.

\bibitem{segal2004machine}
M.~R. Segal, ``Machine learning benchmarks and random forest regression,'' 2004.

\bibitem{wang2021embedding}
L.~Wang, C.~Gao, C.~Huang, R.~Liu, W.~Ma, and S.~Vosoughi, ``Embedding heterogeneous networks into hyperbolic space without meta-path,'' in \emph{Proceedings of the AAAI conference on artificial intelligence}, vol.~35, no.~11, 2021, pp. 10\,147--10\,155.

\bibitem{chang2022megnn}
Y.~Chang, C.~Chen, W.~Hu, Z.~Zheng, X.~Zhou, and S.~Chen, ``Megnn: Meta-path extracted graph neural network for heterogeneous graph representation learning,'' \emph{Knowledge-Based Systems}, vol. 235, p. 107611, 2022.

\bibitem{yun2019graph}
S.~Yun, M.~Jeong, R.~Kim, J.~Kang, and H.~J. Kim, ``Graph transformer networks,'' \emph{Advances in neural information processing systems}, vol.~32, 2019.

\end{thebibliography}

\begin{IEEEbiography}[{\includegraphics[width=1in,height=1.25in,clip,keepaspectratio]{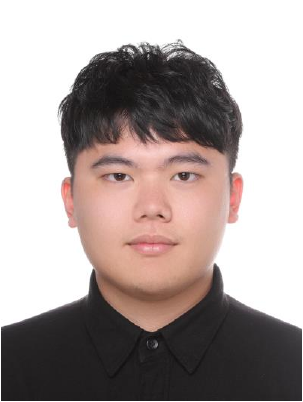}}]{Chengxiang Jin}
	received the BS degree in electrical engineering and automation from Zhejiang University of Science and Technology, Hangzhou, China, in 2021. He is currently pursuing the MS degree in control theory and engineering at Zhejiang University of Technology, Hangzhou, China. His current research interests include graph data mining and blockchain data analytics, especially for heterogeneous graph mining in Ethereum.
\end{IEEEbiography}
\vspace{-35pt}

\begin{IEEEbiography}[{\includegraphics[width=1in,height=1.25in,clip,keepaspectratio]{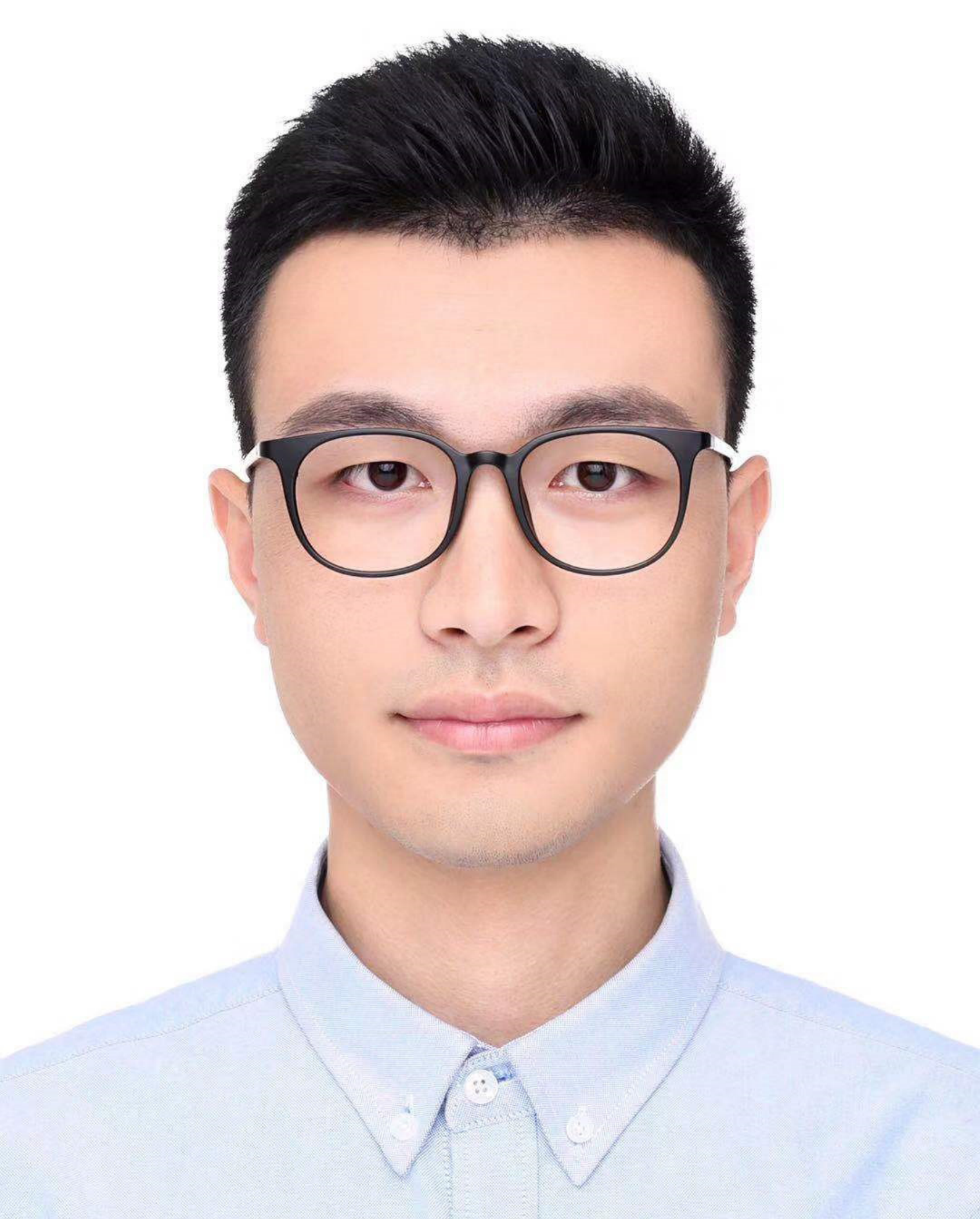}}]{Jiajun Zhou}
	received the Ph.D degree in control theory and engineering from Zhejiang University of Technology, Hangzhou, China, in 2023. He is currently a Postdoctoral Research Fellow with the Institute of Cyberspace Security, Zhejiang University of Technology. His current research interests include graph data mining, cyberspace security and data management.
\end{IEEEbiography}
\vspace{-35pt}

\begin{IEEEbiography}[{\includegraphics[width=1in,height=1.25in,clip,keepaspectratio]{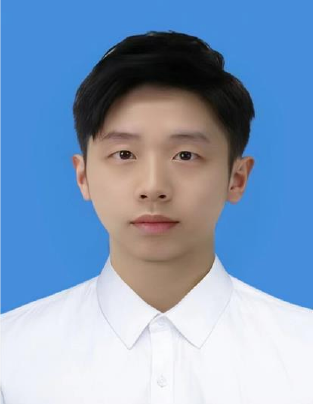}}]{Jie Jin}
	received the BS degree in automation from Hunan International Economics University, Hunan, China, in 2020. He is currently pursuing the MS degree at the College of Information Engineering, Zhejiang University of Technology, Hangzhou, China. His current research interests include Graph data mining and Blockchain Data Analyse.
\end{IEEEbiography}
\vspace{-35pt}

\begin{IEEEbiography}[{\includegraphics[width=1in,height=1.25in,clip,keepaspectratio]{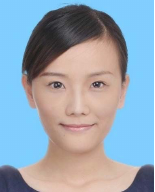}}]{Jiajing Wu}
	(Senior Member, IEEE) received the Ph.D. degree from The Hong Kong Polytechnic University, Hong Kong, in 2014. In 2015, she joined Sun Yat-sen University, Guangzhou, China, where she is currently an Associate Professor. Her research interests include blockchain, graph mining, and network science.Dr. Wu was awarded the Hong Kong Ph.D. Fellowship Scheme during her Ph.D. degree in Hong Kong from 2010 to 2014. She also serves as an Associate Editor for IEEE TRANSACTIONS ON CIRCUITS AND SYSTEMS II: EXPRESS BRIEFS.
\end{IEEEbiography}
\vspace{-35pt}

\begin{IEEEbiography}[{\includegraphics[width=1in,height=1.25in,clip,keepaspectratio]{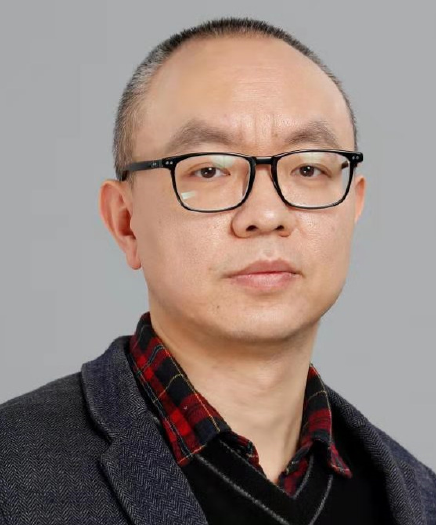}}]{Qi Xuan}(M'18) received the BS and PhD degrees in control theory and engineering from Zhejiang University, Hangzhou, China, in 2003 and 2008, respectively. He was a Post-Doctoral Researcher with the Department of Information Science and Electronic Engineering, Zhejiang University, from 2008 to 2010, respectively, and a Research Assistant with the Department of Electronic Engineering, City University of Hong Kong, Hong Kong, in 2010 and 2017. From 2012 to 2014, he was a Post-Doctoral Fellow with the Department of Computer Science, University of California at Davis, CA, USA. He is a senior member of the IEEE and is currently a Professor with the Institute of Cyberspace Security, College of Information Engineering, Zhejiang University of Technology, Hangzhou, China. His current research interests include network science, graph data mining, cyberspace security, machine learning, and computer vision.
\end{IEEEbiography}

\end{document}